\documentclass{article} % For LaTeX2e
\usepackage{iclr2026_conference,times}
\usepackage{booktabs}
\usepackage{colortbl}
\usepackage{xcolor}
\usepackage{multirow}
\usepackage{fancyhdr}

\usepackage{amsmath,amsfonts,bm}

\def\eqref#1{equation~\ref{#1}}
\def\Eqref#1{Equation~\ref{#1}}
\def\1{\bm{1}}

\def\vc{{\bm{c}}}

\def\vx{{\bm{x}}}

\def\vz{{\bm{z}}}

\def\mA{{\bm{A}}}
\def\mB{{\bm{B}}}

\def\mW{{\bm{W}}}
\def\mX{{\bm{X}}}
\def\mY{{\bm{Y}}}
\def\mZ{{\bm{Z}}}

\DeclareMathAlphabet{\mathsfit}{\encodingdefault}{\sfdefault}{m}{sl}
\SetMathAlphabet{\mathsfit}{bold}{\encodingdefault}{\sfdefault}{bx}{n}

\def\sC{{\mathbb{C}}}
\def\sD{{\mathbb{D}}}
\newcommand{\E}{\mathbb{E}}
\newcommand{\Ls}{\mathcal{L}}
\newcommand{\R}{\mathbb{R}}

\DeclareMathOperator*{\argmin}{arg\,min}

\usepackage{hyperref}
\usepackage{url}
\usepackage{graphicx}
\usepackage{mathtools}
\usepackage[size=small]{todonotes}
\usepackage{cleveref}
\usepackage{blindtext}
\usepackage{caption}
\usepackage{subcaption}
\usepackage{float}
\usepackage{pifont}
\usepackage{multirow}
\usepackage{tcolorbox}

\usepackage{wrapfig}

\usepackage{makecell}

\newcommand{\cmark}{\ding{51}}%
\newcommand{\xmark}{\ding{55}}%

\newcommand{\se}[1]{\textcolor{red}{#1}}

\definecolor{cf}{rgb}{0.63, 0.79, 0.95}

\definecolor{Brown}{rgb}{0.59, 0.29, 0.0}

\graphicspath{{figures/}}

\iclrfinalcopy
\title{Zhyper: Factorized Hypernetworks \\ for Conditioned LLM Fine-Tuning}

\author{Mohamed Hesham Ibrahim Abdalla \\
Department of Computer Science \\
University of Technology Nuremberg \\
\texttt{hesham.abdalla@utn.de}
\And
Zhipin Wang \\
Department of Computer Science \\
University of Technology Nuremberg \\
\texttt{zhipin.wang@utn.de}
\And
Christian Frey \\
Department of Computer Science \\
University of Technology Nuremberg \\
\texttt{christian.frey@utn.de}
\And
Steffen Eger \\
Department of Computer Science \\
University of Technology Nuremberg \\
\texttt{steffen.eger@utn.de}
\\[0.2em]
\and
\makebox[\textwidth][c]{%
\begin{minipage}{0.4\textwidth}
\raggedright
\textbf{Josif Grabocka} \\
Department of Computer Science \\
University of Technology Nuremberg \\
\texttt{josif.grabocka@utn.de}
\end{minipage}}
}

\newcommand{\ourmethod}{Zhyper}

\begin{document}
\maketitle

\pagestyle{fancy}
\lhead{Preprint. Under review.} % change the left header
\chead{} % clear the center header

\begin{abstract}
%Large Language Models (LLMs) have become increasingly popular but often struggle to adapt to new tasks. Standard fine-tuning methods such as Low-Rank Adaptation (LoRA) require carefully curated datasets and significant expertise. Recent work explores using hypernetworks conditioned on textual task descriptions to generate LoRA adapters on the fly, enabling task adaptation without task-specific fine-tuning, but these approaches are parameter-intensive. We propose \textbf{Zhyper}, a parameter-efficient factorized hypernetwork framework that generates context-aware LoRA adapters from textual descriptions. Zhyper drastically reduces inference-time parameters while preserving flexibility across diverse tasks. Experiments on multiple benchmarks show that Zhyper achieves competitive performance with up to \textbf{20x} fewer parameters than state-of-the-art baseline. Furthermore, we extend Zhyper to cultural alignment, demonstrating improved generalization to out-of-domain settings by capturing fine-grained contextual values more effectively than existing frameworks. These results establish hypernetwork-conditioned adaptation as a scalable and efficient path toward building adaptable and value-sensitive LLMs.

Large Language Model (LLM) conditioning refers to instructing an LLM to generate content in accordance with the norms and values of a specific culture, beliefs of a particular political orientation, or any desired text-specified semantic conditioning. Unfortunately, prompt engineering does not ensure that LLMs behave in accordance with a desired conditioning due to the inductive bias of the pre-training and alignment datasets. Prior works have focused on fine-tuning LLMs by directly conditioning the LoRA weights; however, such methods introduce a large number of parameters. As a remedy, we propose Zhyper, a parameter-efficient factori\textbf{z}ed \textbf{hyper}network framework that generates context-aware LoRA adapters from textual descriptions. Experiments on multiple benchmarks show that Zhyper achieves competitive performance with up to \textbf{26x} fewer parameters than the state-of-the-art baselines. Furthermore, we extend Zhyper to cultural alignment, demonstrating improved generalization to out-of-domain settings and a better capturing of fine-grained contextual values. 

\end{abstract}

\section{Introduction}
\label{sec:introduction}

%\todo[inline]{SE: we can think of a more technical or a broader, less technical but more ethics-oriented introduction focusing on the big picture}

%%%
%%% OLD PAR- outcommented
%%%
\iffalse
\se{Large Language Models (LLMs) have transformed Natural Language Processing (NLP), Computer Vision (CV), and Machine Learning (ML) more broadly. They achieve state-of-the-art performance in text generation and comprehension across diverse domains, including code synthesis \citep{roziere2023code}, mathematical reasoning \citep{ahn-etal-2024-large}, scientific writing \citep{geng-etal-2025-impact,eger2025transformingsciencelargelanguage}, multimodal tasks such as text–image understanding and generation \citep{alayrac2022flamingo}, and %even 
evaluation of machine translation and related tasks \citep{gu2025surveyllmasajudge}. Despite these advances, LLMs continue to exhibit significant limitations \citep{kostikova2025lllmsdatadrivensurveyevolving}, including hallucinations \citep{kalai2025languagemodelshallucinate}, reasoning errors, and more. One particularly pressing concern is the issue of \emph{value alignment} \citep{alkhamissi-etal-2024-investigating}: ensuring that LLMs produce outputs consistent with specific human values and ethical principles. Without careful alignment, LLMs risk reproducing only the cultural assumptions embedded in their predominantly English-centric training data or reflecting the perspectives of their (often US- or China-based) developers \citep{buyl2025largelanguagemodelsreflect}.}
\fi
%%%%
%%%% END OLD PAR
%%%%

% MOTIVATION / BACKGROUND
Large Language Models (LLMs) have transformed Natural Language Processing (NLP), Computer Vision (CV), and machine learning (ML) more broadly. They achieve state-of-the-art performance in text generation and comprehension across diverse domains, including code synthesis \citep{roziere2023code}, mathematical reasoning \citep{ahn-etal-2024-large}, scientific writing \citep{geng-etal-2025-impact,eger2025transformingsciencelargelanguage}, multimodal tasks such as text–image understanding and generation \citep{alayrac2022flamingo}, and %even 
evaluation of machine translation and related tasks \citep{gu2025surveyllmasajudge}. 
This success stems from scaling to millions and billions of parameters. However, this scaling requires large computational resources, motivating the search for parameter-efficient fine-tuning (PEFT) techniques.

% CURRENT SOTA / LIMITATIONS
Recent advances have made it possible to adapt LLMs 
to task-specific criteria, which is crucial for a broader applicability and acceptance of NLP systems. 
A recent stream of research leverages PEFT techniques \citep{Ding_2023_Parameter-efficient, Weyssow2023Exploring, Prottasha2024Parameter-efficient}, e.g., Low-Rank Adaptions (LoRA) \citep{hu_2021_lora} to adapt for desired task-specific values in an LLM. 
LoRA achieves this by freezing most of the pre-trained model's parameters and introducing trainable low-rank matrices, yielding weight correction terms. However, stand-alone LoRA approaches are primarily tailored for a single-task adaptation and may lose their effectiveness in a setting where an LLM needs to be adapted to various downstream settings. Therefore, approaches directly tackling a multi-task learning (MTL) setting have been proposed \citep{agiza2024mtlora,wang2023multilora,luo2024moelora,wang2024malora} that aim to do multi-task fine-tuning efficiently, where a shared backbone model must serve multiple tasks. 
%An extension leveraging quantization, namely QLoRA, reduces memory consumption, however, shows variability in results across several runs \cite{Alahmari_2024_Repeatability}. 
A promising direction for the dynamic and robust individualization of LLMs is by leveraging \emph{hypernetworks} in the training pipeline. In Text-to-LoRA (T2L) \citep{charakorn_2025_text2lora}, the authors apply hypernetworks to adapt LLMs to specific task descriptions using only a textual task description as the input for learning the adapters' weights. 
However, two open challenges remain unresolved. First, existing conditioned LoRA methods, such as T2L, are not parameter-efficient when extended to large contextual spaces. Second, the applicability of conditioned LoRA tuning has not been explored for the important real-world problem of cultural alignment.

\begin{figure}
    \centering \includegraphics[width=0.9\linewidth]{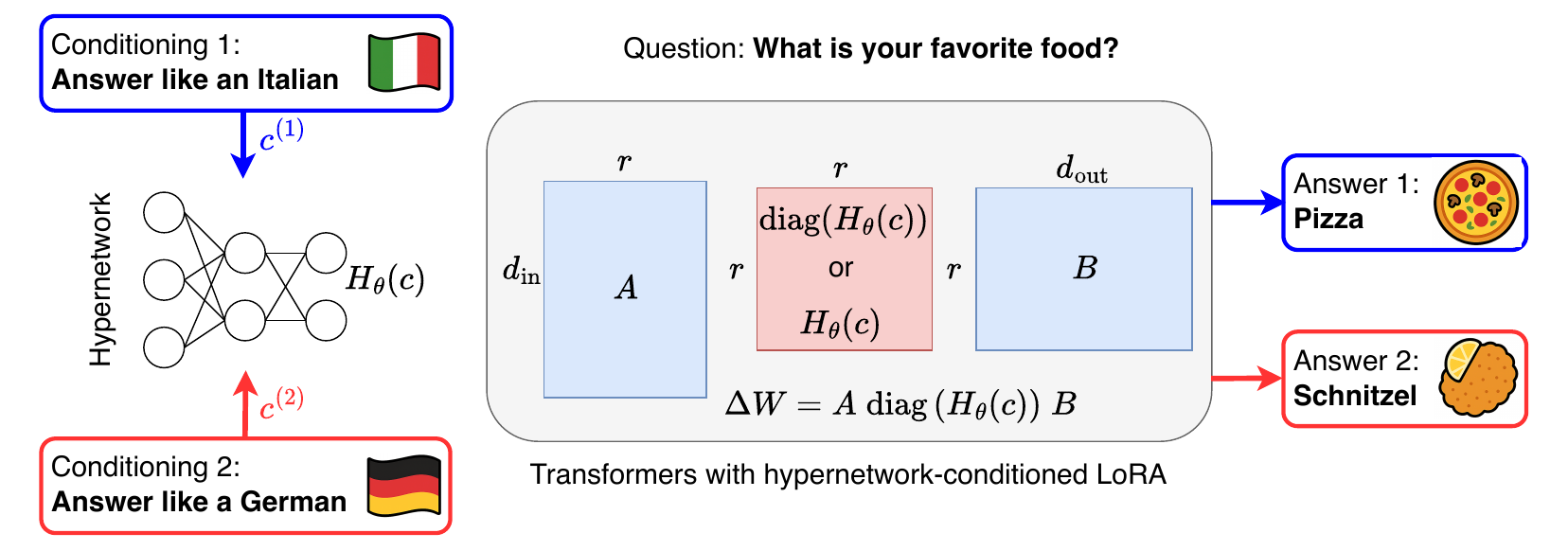}
    \caption{We introduce a novel parameter-efficient architecture for conditioned LLM finetuning based on hypernetwork-conditioned LoRA adapters}
    \label{fig:workflowExample}
\end{figure}

To tackle the described challenges, we propose a factori\textbf{z}ed \textbf{hyper}network, called \textbf{\ourmethod}, which leverages a hypernetwork to inject desired values in the outputs of an LLM. More specifically, the hypernetwork should produce a different weight based on the current layer, a layer type for attention-awareness, and the respective description of a context we want to adapt to. As opposed to prior works \citep{charakorn_2025_text2lora}, we additionally experiment with contexts being descriptions of cultures. Considering the example shown in \Cref{fig:workflowExample}, the goal is to condition a base model on certain criteria. For instance, when choosing a preferred food, the answer might have country-specific dependence. The contextual modulation signal is computed via a hypernetwork that is integrated into the computation of the LoRA adapter, leading to answers conditioned on the instilled values. 

% insights in results
We empirically show that our novel model achieves comparable predictive performance at an order of magnitude fewer parameters on a variety of LLM capability assessments, e.g., math, science, coding, reasoning, and word knowledge. Furthermore, we provide a thorough ablation study on the contextual modulation signal represented as an $(r\times r)$-matrix, where $r$ denotes the rank of the LoRA adapters.  %an alignment category. 

% contribs
Our contributions are as follows:
\begin{itemize}
    \item A novel lightweight hypernetwork-based architecture for training LoRA adapters that align to text 
    or culture descriptions with up to \textbf{26x} fewer parameters compared to prior work.
    \item Hypernetwork that generates a compact contextual modulation signal instead of generating all parameters of an adapter.
    \item A thorough empirical study on efficient learning strategies for the conditioned fine-tuning of Large Language Models.
    \item Improved empirical performances in the important use cases of task conditioning and cultural alignment.
\end{itemize}

\section{\ourmethod~- Conditioned LLM Tuning}
\label{sec:method}
Our method \ourmethod~leverages hypernetworks to induce descriptive information and generate LoRA adapters for context-specific adaptation. The following subsections present the preliminaries in \Cref{subsec:preliminaries}, our novel factorized architecture in \Cref{subsec:architecture}, and the complexity analysis of our method in \Cref{subsec:complexity}.

%\todo[inline]{SE: we may want a similar method section for the evaluation part}
%\todo[inline, color=cf]{CF: not sure what you mean here. Do you mean to move the metrics to a separate section?}

\subsection{Preliminaries.} 
\label{subsec:preliminaries}
\textbf{Low-Rank Adaptation (LoRA)} is a well-established parameter-efficient fine-tuning technique for LLMs~\citep{hu_2021_lora}. Generally, the weights of a base model are frozen, and only low-rank weight matrices are trained, serving as adapters to the model. Formally, for each selected linear transformation $h=\mW^{\text{base}} \vx$, the fine-tuned transformation is given by $h'=\mW^{\text{base}} \vx + \Delta \mW \vx$, 
%\todo{SE: this is confusing --- $h$ has already been defined as something else --- maybe you mean $h'$?}
with $\Delta \mW = \mA \mB $, where $\mA \in \R^{d_{\text{in}} \times r}$, and $\mB \in \R^{r \times d_{\text{out}}}$ are low-rank weight matrices with $r \ll d$.

\textbf{Hypernetworks} introduce 
neural networks whose output defines the parameters of another network ~\citep{ha2016hypernetworks}. It 
formalizes the idea of learning to generate weights rather than learning weights directly. Formally, let $f_\theta(x)$ denote a parameterized target network with $\theta \in \R^{r}$. A parameterized hypernetwork $H_\phi(z): \R^{m} \rightarrow \R^{r}$ is conditioned on a layer-specific descriptor vector $\vz \in \R^{m}$ and parameters $\phi$, and outputs the weights $\theta$ of the target network.

\subsection{Architecture}
\label{subsec:architecture}
We present the \textbf{\ourmethod} method, a hypernetwork-conditioned low-rank adaptation method that enables parameter-efficient and context-aware fine-tuning of LLMs. The general workflow of our method is illustrated in \Cref{fig:workflowExample}, where in the following we provide details on the respective components. 

\textbf{Contextual Information.~}We represent contextual features (e.g., value or cultural descriptions) leveraging a transformer-based encoder trained for general text embeddings. Each description is transformed into a fixed-length embedding vector $\vc \in \R^{d_{c}}$, which serves as the contextual input to our hypernetwork described below. This representation ensures that diverse textual descriptions are mapped into a unified semantic space suitable for conditioning LoRA adapters. We denote by $\vc_i$ the contextual information associated with the $i$-th dataset.

%\textbf{Hypernetwork-conditioned LoRA.~}
\textbf{Factorized Hypernetworks (\ourmethod-diag).} 
Let $\sD=\{D_i\}_{i=1}^n$ be fine-tuning datasets, where $D_i=\{(\mX_i, \mY_i)\}$ is a set of input-label pairs. Each dataset $i$ is associated with a set of contextual descriptions $\sC _i \coloneqq \{\vc_i^{(j)}\}_{j=1}^M$ where $\vc_i^{(j)} \in \R^{d_c}$. During training, we sample $D_i \sim \sD$ and $\vc_i \sim \sC_i$. 
%\todo{SE: shouldn't that be bold?}
%\todo[color=cf]{CF: yes, thx}

%For training our model, we utilize multiple fine-tuning datasets $\sD=\{D_1, \ldots, D_n\}$.
%For aligning to contextual information, we define $M$ %many descriptor variables
%contextual vector representations $\sC _i \coloneqq \{\vc_i^{(j)}\}_{j=1}^M$ where $\vc_i^{(j)} \in \R^{d_c}$ 
%\todo{SE: what is $e$? $i$?}
%for the $i$-th dataset. 
%Therefore, each dataset is associated with a set of contextual information, i.e.,  $D_i = (\vx_i, \vy_i, \sC_i)$. 
%In the following, let $l\in\{1,\ldots, L\}$ be a layer index of the base model and $t$ denote %s 
%the target module, i.e., the query $\mQ$ and value $\mV$ projection in every attention block of the base LLM.
For each transformer layer $ \ell\in\{1,\ldots, L\}$ and attention projection $t \in \{Q, V\}$ of the base LLM, we learn module-type and layer-specific embeddings. 
%for the calculation of the query or key values. 
For that, we utilize learnable embeddings $e_t=E_{\textit{type}}(t) \in \R^{d_t}$ and  $e_\ell = E_{\textit{layer}}(\ell) \in \R^{d_\ell}$, shared across training. 
Our hypernetwork $H_\phi: \R^{d_c+d_t+d_\ell} \rightarrow \R^r$ is defined to map the concatenated input to a rank-$r$ vector: 
%\begin{equation}
%H_\phi(\vc_i^{(j)}~\|~E_{type}[t]~\|~E_{layer}[l]) = \vz^i_{l, t},
%\end{equation}
\begin{equation}
\vz^i_{\ell, t} = H_\phi(\vc_i^{(j)}~\|~e_t~\|~e_\ell)
\end{equation}
\noindent%
where $\|$ denotes the concatenation operator. Intuitively, $\vz^i_{\ell, t} \in \R^{r}$ denotes a latent representation of a contextual encoding for the $i$-th dataset w.r.t.\ the $\ell$-th layer and the attention component $t$, i.e., query or value projections. 
This leads to the following update rule for the base model's weights: 
%We define the transformation of the base model weights as follows:
\begin{align}
    \Delta \mW_{\ell,t}(c) &= \mA_{\ell,t}  \operatorname{diag}(\vz^i_{\ell,t}) \mB_{\ell,t} \quad \text{with} \quad \mA_{\ell,t}\in \R^{d_{\text{in}} \times r}, \mB_{\ell,t} \in \R^{r \times d_{\text{out}}} \label{eq:deltaTerm}\\
    \mW_{\ell,t}^{\text{adapt}} \vx &\leftarrow (\mW_{\ell,t}^{\text{base}} + \Delta \mW_{\ell,t})
    \vx %= \mW_0 \vx + \mB^T\operatorname{diag}(\vz)~\mA \vx,
\end{align}
%\begin{equation}
%    \mW_0 \vx + \Delta \mW \vx = \mW_0 \vx + \mB^T\operatorname{diag}(\vz)~\mA \vx,
%\end{equation}
where $\operatorname{diag}(\vz^i_{\ell,t}) \in \R^{r \times r}$ yields a diagonal matrix with the elements of $\vz^i_{\ell,t}$ on the diagonal.

The \textbf{{\ourmethod}-square} variant is an ablation of our method where the hypernetwork outputs a square matrix $H_\phi: \R^{d_c+d_t+d_\ell} \rightarrow \R^{r \times r}$, leading to $\Delta \mW_{\ell,t}(c) = \mA_{\ell,t}  \vz^i_{\ell,t} \mB_{\ell,t}$ where $\vz^i_{\ell,t} \in R^{r \times r}$.

\textbf{Training Objective.} To integrate the hypernetwork-generated LoRA adapters into the base model with weights $\mW^{\text{base}}$, we formalize the training objective as minimizing the supervised fine-tuning loss over datasets and their associated contextual descriptors, ensuring that each layer and module type is conditioned on context-specific information. We define the trainable parameters $\theta=\{A_{\ell,t}, B_{\ell,t}, \phi, E_{\textit{type}}, E_{\textit{layer}}\}$. The supervised fine-tuning training objective becomes:
\begin{equation}
    %\theta_{t,l} \leftarrow 
    \underset{\theta}{\argmin}~\E_{i \sim [n]}\E_{(x,y) \sim \sD_i}\E_{c_i^{(j)} \sim \sC_i} \; \Ls_{\text{SFT}} 
    \left(f_{W^{\text{base}},\Delta \mW\left(c_i^{(j)}\right)}(x)\: ,\: y\right)%(D_i, \mW^{\text{base}}, H_\phi(c_i^{(j)}~\|~e_t~\|~e_l)),
\end{equation}
where $f_{W^{\text{base}},\Delta \mW(c_i^{(j)})}$ denotes our model's output given the frozen weights of the base model $\mW^{\text{base}}$ and $\Delta \mW(c_i^{(j)})$ denoting the adaptation according to \Cref{eq:deltaTerm}. 

The architecture of our framework enables to train the matrices $\mA$ and $\mB$ once, whereas the hypernetwork provides an efficient contextual modulation by either providing a diagonal scaling matrix or a full square matrix. By factorizing conditioning into the modulation signal we achieve high efficiency while enabling context-aware alignment.
\subsection{Complexity Analysis}
\label{subsec:complexity}
We provide a complexity analysis of our approach compared to our competitors T2L \citep{charakorn_2025_text2lora} and HyperLoRA \citep{lv_2024_hyperlora}, leveraging hypernetworks with respect to the per-context materialization, their representativeness, and generalization capabilities.

\textbf{Per-Context Materialization.} For a transformer with $L$ layers and attention projections $t\in \mathcal{T}$ (e.g., $Q$, $V$), each linear map is adapted by a rank-$r$ LoRA adapter. Let $P_{\ell,t} \coloneqq r(d_{\text{in}} + d_{\text{out}})$ be the number of LoRA parameters per $(\ell,t)$-pair. The hypernetwork parameters are denoted by $P_H$. 

% and the number of distinct context used at inference by $M$.
The hypernetwork's output size is given as $\sum_{l,t} P_{l,t}$ for HyperLoRA~\citep{lv_2024_hyperlora} and T2L~\citep{charakorn_2025_text2lora}. Regarding \ourmethod , it is $\sum_{\ell,t} r $ or $\sum_{l,t} r^2 $ depending on the configuration -diag or -mix, respectively. In practical scenarios, we have that $r \ll d_{\text{in}},d_{\text{out}}$, hence, $r^2 \ll r(d_{in} + d_{out})$. Therefore, in \textbf{inference}, both variants of \ourmethod~are far lighter than HyperLoRA and T2L. The per-context GPU memory scales as:
\begin{align}
\text{\{HyperLoRA , T2L\}} \gg \text{\ourmethod-square} \geq \text{\ourmethod-diag}
\end{align}
In terms of \textbf{trainable parameters}, HyperLoRA trains $P_H+P_{\text{emb}}$ parameters, where $P_{\text{emb}}$ refers to their task query embeddings. Similarly, T2L trains on $P_H+P_{\text{layer}}(L, d_e) + P_{\text{type}}(\mathcal{T}, D_e)+ P_{\text{emb}}$ parameters, i.e., layer- and type-wise embeddings are added. The learnable parameters of \ourmethod~ aggregates to $\sum_{\ell,t} P_{\ell,t} + P_H + P_{layer}(L, d_e) + P_{type}(\mathcal{T}, D_e)$. For the models HyprLoRA and T2L, $P_H$ has to be sufficiently large such that $(A,B)$ matrices of the LoRA adapters can be generated with high fidelity. In \ourmethod, we follow the idea of paying $\sum_{\ell,t} P_{\ell,t}$ once, and the hypernetwork outputs only rank-$r$ matrices as modulation signals. Therefore, in our method, $P_H$ is much smaller compared to T2L and HyperLoRA, where the hypernetwork emits $(A,B)$ directly. 

\textbf{Representativeness.} Let $\mathcal{H}_{\text{full}} = \{\mA\mB: \mA \in \R^{d_{\text{in}} \times r}, \mB \in \R^{r \times d_{\text{out}}}\}$ be the hypothesis class of a LoRA adapters. 
That is, HyperLoRA and T2L can realize any element of $\mathcal{H}_{\text{full}}$ subject to their hypernetwork's capacity. 
For \ourmethod-diag, we define $\mathcal{H}_{\text{diag}} = \{\mA \operatorname{diag}(\vz) \mB: \mA \in \R^{d_{in} \times r}, \mB \in \R^{r \times d_{out}}, \vz \in \R^r\}$ that defines a strict subset of $\mathcal{H}_{\text{full}}$. 
Likewise, we define $\mathcal{H}_{\text{square}} = \{\mA \mZ \mB: \mA \in \R^{d_{in} \times r}, \mB \in \R^{r \times d_{out}}, \mZ \in \R^{r \times r}\}$ for which $\mathcal{H}_{\text{square}}$ matches $\mathcal{H}_{\text{full}}$ iff $\mA$ and $\mB$ have full row/column rank $r$. 
Therefore, \ourmethod-full can approximate any adapter in $\mathcal{H}_{\text{full}}$. This leads to the relationship:
\begin{align}
\mathcal{H}_{\text{diag}} \subseteq \mathcal{H}_{\text{square}} \subseteq \mathcal{H}_{\text{full}}
\end{align}

\textbf{Generalization.} Given the hypothesis classes and the number of free parameters for each of model, we have that the Rademacher complexity scales with $\mathfrak{R}(\mathcal{H}_{\text{full}}) = \mathcal{O}\left( \sqrt{\frac{r(d_{\text{in}} + d_{\text{out}})}{N}} \right)$, where $N$ is the sample size \citep{shwartz_2020_bible}. Likewise, we get that $\mathfrak{R}(\mathcal{H}_{\text{diag}}) = \mathcal{O}\left( \sqrt{\frac{r}{N}} \right)$ and $\mathfrak{R}(\mathcal{H}_{\text{square}}) = \mathcal{O}\left( \sqrt{\frac{r^2}{N}} \right) = \mathcal{O}\left( \frac{r}{\sqrt{N}} \right)$.
This leads to the relationship:
\begin{align}
\label{eq:rademacher}
\mathfrak{R}(\mathcal{H}_{\text{diag}}) \leq \mathfrak{R}(\mathcal{H}_{\text{square}}) \leq \mathfrak{R}(\mathcal{H}_{\text{full}})  
\end{align}

By constraining the hypothesis classes that lower the Rademacher complexity, we get tighter generalization bounds for \ourmethod(-diag, or -square) compared to HyperLoRA and T2L. 
Notably, in practical settings with $r \ll (d_{in}+d_{out})$, the inequalities in \Eqref{eq:rademacher}~become strict. Consequently, our model's performance is likely to transfer to unseen data whilst reducing the risk of overfitting and using an order of magnitude fewer parameters compared to other competitors. 
\section{Experiments}
\label{sec:evaluation}

%\todo{SE: to save space, we could use \S for section, but then we should do it consistently}
%\todo[color=cf]{CF: according to the style sheet, the recommendation here is rather to use "section"}

In our experimental protocol, we address two important real-world use cases:

\begin{itemize}
    \item \textbf{Task Conditioning}: where LLMs are conditioned to perform a certain task, e.g., to act as an expert on geography, similar to the setting of T2L \citep{charakorn_2025_text2lora} (Section~\ref{subsubsec:exp_hypotheses_task}).
    \item \textbf{Cultural Alignment}: where LLMs are instructed to generate content aligned with the norms and values of a culture, e.g., to write like a European (Section~\ref{subsubsec:exp_hypotheses_align}).
\end{itemize}
%
%\todo[inline,color=yellow]{Zhipin: I think it would be clearer to move these cultural training and evaluation details before the CulturalBench results section.}

%\todo[inline, color=cf]{CulturalBench, CultureBank, Prism,...\\@Zhipin: do we have results on any of the other benchmarks? Also, write here how you generate the templates for the loras.}

%\textbullet~\emph{Evaluation Metrics.~} 
%\todo[inline, color=cf]{@Zhipin: which metrics do you use in tab2? do we need all of them here. seems like after wvs are not evaluted, some of them can be outcommented for the moment (content of the next paper)}

\textbullet~\emph{Hyperparameters of our method.} We use a 3-layer MLP, with the weight of output head of size $d_{\text{MLP\_out}}\ \times r$ which is different from T2L head, with weight of $d_{\text{MLP\_out}}\ \times r \times (d_{\text{out}} + d_{\text{in}})$ where $d_{\text{MLP\_out}}$ is the output size of the last MLP block. To generate the embeddings of the text descriptions, we use \texttt{gte-large-en-v1.5} \citep{zhang2024mgte, li2023towards}. Our method introduces a new hyperparameter, $Z$ matrix type, which can be either a diagonal matrix or a square matrix. Using this hyperparameter together with the LoRA rank, we conduct a hyperparameter analysis on a subset of the benchmark dataset (validation set). We find that the configuration with $r=8$ and a diagonal $Z$ matrix achieves the best performance 10 task-based benchmark subsets while maintaining a low number of parameters ($\sim$~2.5M). In evaluation, we refer to this variant as simply \textbf{\ourmethod}. We perform a similar hyperparameter tuning procedure for the cultural alignment models. Comparisons between different variants are provided in Appendix Section~\ref{subsec:hyperparam_tuning}.

The source code of our framework and experiments is publicly available.\footnote{\url{https://github.com/machinelearningnuremberg/Zhyper}}
%\todo{SE: that should be in the conclusion or intro maybe}
%\todo{SE: hypotheses may want to come earlier}
%
%\subsection{Results}
%\label{subsec:exp_hypotheses}
%
\subsection{Use case on Task Conditioning}
\label{subsubsec:exp_hypotheses_task}
%\begin{hypothesis}
%Our novel conditioning achieves comparable predictive performance at an order of magnitude less parameters.
%\end{hypothesis}

\textbullet~\emph{Baselines.} 
We evaluate our method on \texttt{\texttt{Mistral-7B-Instruct-v0.2}} \citep{jiang_2023_mistral7b} as an unconditioned baseline model, along with a variant utilizing few-shot in-context learning (ICL)~\citep{brown_2020_languagemodelsfewshotlearners,{dong_2024_ICLsurvey}}, and another that incorporates prepended task descriptions in the query. As fine-tuned models, we compare against T2L (SFT) L \citep{charakorn_2025_text2lora}, which
performs instant adaptation of LLMs from task descriptions; multi-task LoRA (MTL), a variant of LoRA trained on all tasks; task-specific LoRA (Oracle), trained only on the corresponding task; and Hyperdecoders \citep{ivison_2022_hyperdecoders}, which generate LoRAs on a per-sequence basis. We also report the zero-shot results of Arrow Routing \citep{ostapenko_2024_arrow}; because code is unavailable, we copy their reported numbers, which use LoRA rank $r$ of 4. Our experiments show that the best-performing T2L variant uses $r=16$, while the best MTL variant uses $r=8$. For completeness, we also report results for LoRA ranks $r=8$, $r=16$, and $r=32$.
\\\leavevmode\\
\textbullet~\emph{Datasets.} We use the SNI dataset \citep{Wang_2022_sni} to \textbf{train} our task-based model. Following the T2L setup, 11 tasks are held out for evaluation, and 10 datasets are removed to avoid data contamination with the evaluation benchmarks, leaving 479 datasets for training. We also reuse the task descriptions generated in T2L, with 128 descriptions per training dataset. For \textbf{evaluation}, we utilize 10 benchmark datasets that enable a broad assessment across diverse areas, such as reasoning, math, science, coding, and world knowledge. We evaluate on the following benchmarks: 
Arc-challenge (ArC) and Arc-easy (ArE) \citep{Clark_2018_Arc}, 
OpenBookQA (OQA) \citep{mihaylov_2018_openbook}, 
HumanEval (HE) \citep{Chen_2021_humaneval}, 
HellaSwag (HS) \citep{Zellers_2019_HellaSwag}, 
MBPP \citep{austin_2021_mbpp}, 
Winogrande (WG) \citep{keisuke_2021_winogrande}, 
GSM8K \citep{Cobbe_2021_TrainingVT}, 
PIQA \citep{Bisk_2019_PIQA}, and
Boolq (BQ) \citep{clark_2019_boolq}. These benchmarks are excluded from training unless explicitly used as an oracle, and are therefore treated as unseen. Each benchmark is evaluated using three different text descriptions, and the results are averaged across them.

\begin{table}[t]
\centering
\caption{Benchmark performance on unseen tasks and descriptions. T2L, MTL and Task-specific LoRAs results are reproduced by us, while the others are taken from T2L \citep{charakorn_2025_text2lora}. All methods use a LoRA rank of $r = 8$, except for Arrow Routing, which uses $r = 4$ and T2L with $r = 16$. Best numbers per column are in \textbf{bold}.}
\resizebox{\columnwidth}{!}{%
\renewcommand{\arraystretch}{1.2}
\begin{tabular}{lccccccccccc|c}
\toprule
 & \multirow{2}{5em}{\textbf{Trainable Params}} & \textbf{ArcC} & \textbf{ArcE} & \textbf{BQ} & \textbf{HS} & \textbf{OQA} & \textbf{PIQA} & \textbf{WG} & \textbf{MBPP}  & \textbf{GSM8K} & \textbf{HE} & \textbf{Avg.} \\
 & & (acc) & (acc) & (acc) & (acc) & (acc) & (acc) & (acc) & (pass@1) & (acc) & (pass@1) & (10 tasks) \\
\midrule
\multicolumn{12}{l}{\textbf{Zero-shot adaptation without fine-tuning}} \\
Mistral-7B-Instruct & N/A & 65.4 & 77.8 & 71.6 & 49.7 & 54.2 & 72.8 & 45.0 & 43.1 & 40.9 & 37.2 & 55.8 \\
Prepending task desc. & N/A & 72.0 & 85.8 & 67.6 & 58.9 & 63.4 & 77.9 & 59.0 & 41.6 & 40.9 & 39.0 & 60.6 \\
\midrule
\multicolumn{13}{l}{\textbf{Few-shot adaptation without fine-tuning}} \\
3-shot ICL & N/A & 72.1 & 85.9 & 71.7 & 59.0 & 66.2 & 76.2 & 58.0 & 42.6 & 40.9 & 37.2 & 61.0 \\
\midrule
\multicolumn{12}{l}{\textbf{Zero-shot adaptation after fine-tuning}} \\
Arrow Routing $(r=4)$ & N/A & 60.9 & 86.2 &  87.6 &  80.8 & 48.6 & 83.0 & \bf 68.5 & 50.2 & N/A & 28.7 & N/A \\
Hyperdecoders & 55M &  76.6 & 88.5 & 83.9 & 65.2 &  76.6 & 81.3 &  64.9 & 51.6 & 43.6 &40.9 &  67.3 \\
MTL & 3.4M & 74.0 & 87.3 & 84.0 & 63.4 & 69.2 &  81.5 & 60.5 & 49.1 & 47.5 &  39.6 & 65.4 \\
\midrule
\multicolumn{13}{l}{\textbf{Fine-tuned directly on test tasks (Oracle)}} \\
Task-specific LoRAs & 3.4M &  74.6 &  88.3 & \bf 88.0 & \bf 87.9 & \bf 77.4 & \bf 86.1 & 57.0 & 47.9 & \bf 50.2 & N/A & N/A \\
\midrule
\multicolumn{13}{l}{\textbf{Conditioned zero-shot adaptation after fine-tuning}} \\
%\midrule
%\multicolumn{12}{l}{\textbf{Fine-Tuning Adaptation}} \\
% T2L (SFT) L & 55M & 76.2 & \bf 88.8 & 84.5 & 65.5 & 72.1 & 81.1 & 61.0 & 49.8 &  47.9 & 38.2 &  66.5 \\
T2L (SFT) L $(r=16)$ & 110M & 74.5	& \bf 87.7	&85.5	&64.9	&68.7	&81.5	&59.8	&52.4	&46.5	& \bf42.3	&66.4 \\
\emph{\ourmethod~(Ours)} &  4.2M &  \bf 74.7 & 87.2 &  85.4 & 66.0 & 68.6 & 81.0 & 59.3 & \bf 52.6 & 44.2 &  39.6 & 65.9 \\
\bottomrule
\end{tabular}
}
\label{tab:t2lvsours}
\end{table}

\begin{wraptable}{r}{0.5\textwidth}
    \centering
    \caption{Number of parameters.}
    \resizebox{0.5\textwidth}{!}{
    \begin{tabular}{c r r r r}
    \toprule
         \thead{LoRA Rank} & \thead{\makecell{MTL}} & \thead{\makecell{Zhyper-diag}} & \thead{\makecell{Zhyper-square}} & %\thead{\makecell{Hypernetwork outputs}}\\
         \thead{\makecell{T2L}}\\
         \midrule
         %1 & 0.43M & 1.22M & 1.22M & 7.62M \\
         %4 & 1.70M & 2.50M & 2.51M & 28.11M \\
         8 & 3.41M & 4.21M & 4.27M & 55.00M\\
         16 & 6.82M & 7.62M & 7.87M & 110.06M \\
         32 & 13.63M & 14.46M & 15.47M & 219.32M \\
         \midrule
         \textbf{Avg. Performance} & 64.0 & 64.8 & 64.3 & 65.6 \\
         \bottomrule
    \end{tabular}
    }
    \label{tab:num_params_exact_comaprison}
\end{wraptable}

\begin{figure}[h]
    \centering
    \hspace{0.75cm}
    \begin{subfigure}[b]{0.345\linewidth}
        \centering
        \includegraphics[width=\linewidth]{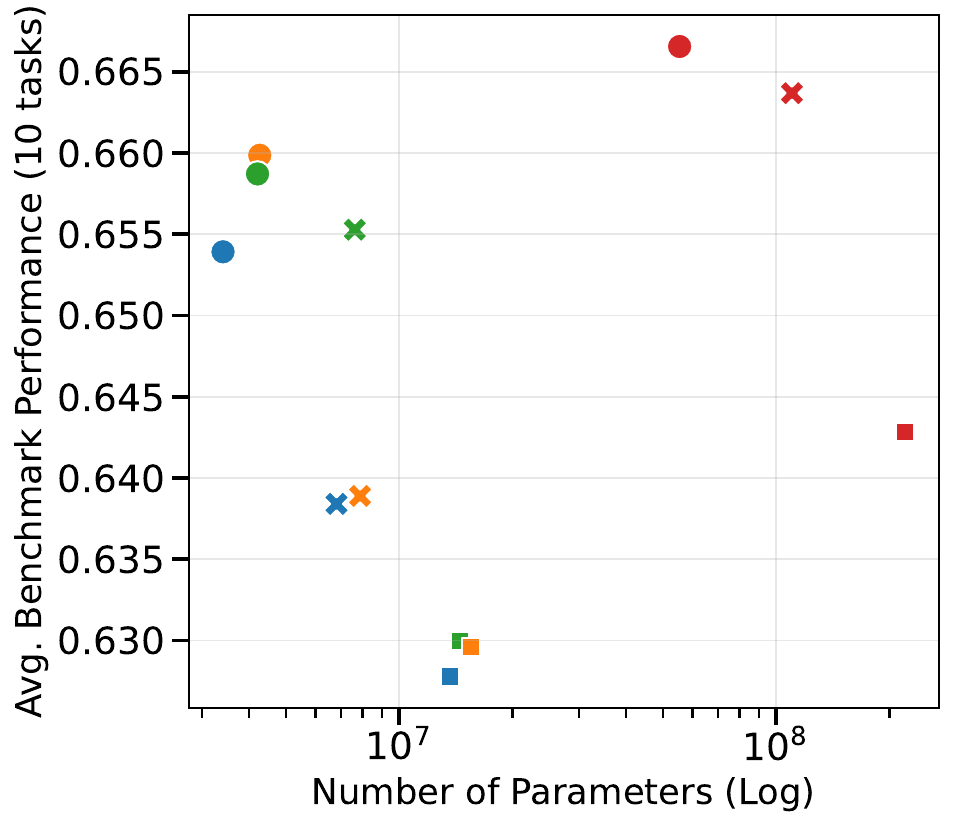}
    \end{subfigure}
    \begin{subfigure}[b]{0.475\linewidth}
        \centering
        \includegraphics[width=\linewidth]{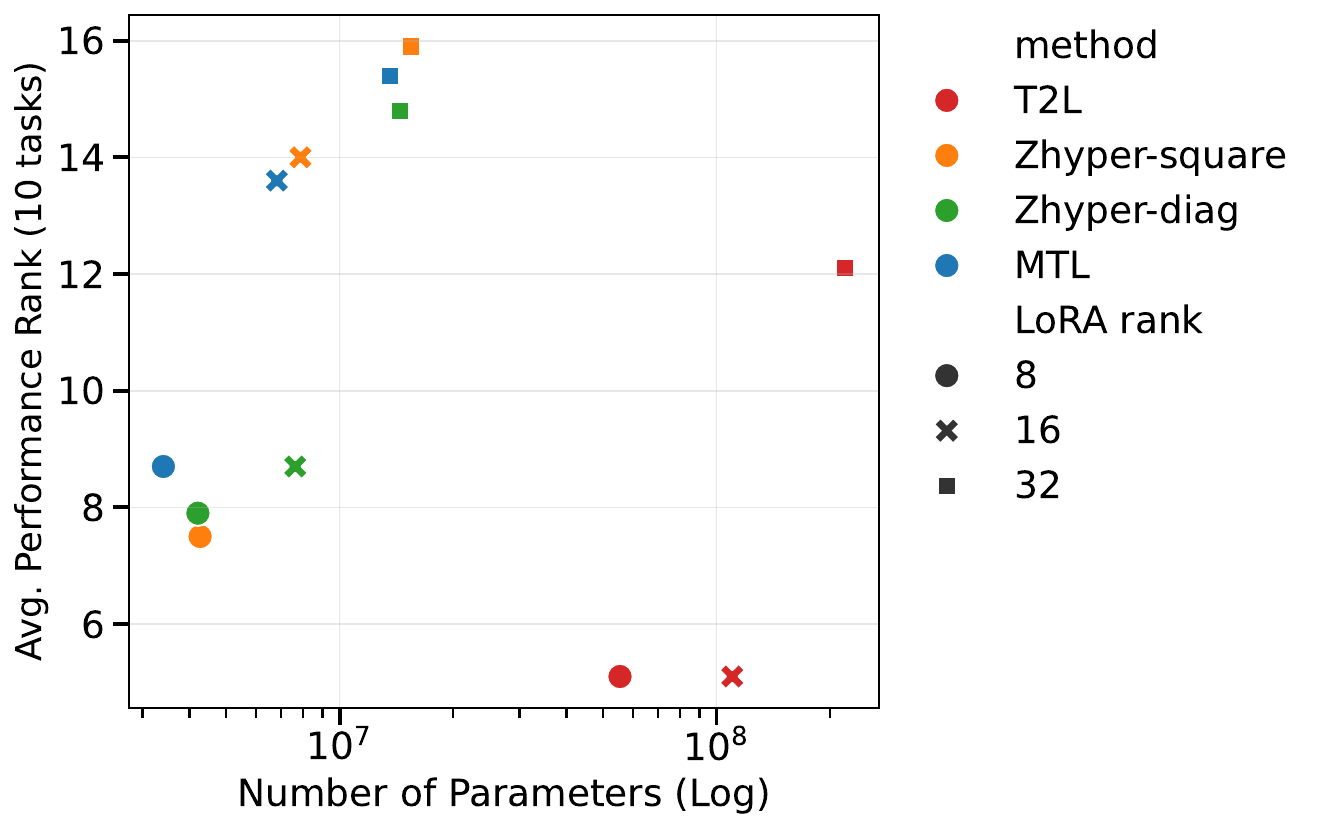}
    \end{subfigure}
    \caption{%Left: Number of parameters vs. performance metrics. 
    (Left) Average performance (higher is better); (Right) Performance rank (lower is better). Our method lies in the Pareto front optimality between performance and the number of parameters. }
    \label{fig:ours_vs_baseline_scatter}
\end{figure}

\begin{figure}[t]
    \centering
%    \begin{subfigure}[b]{0.49\linewidth}
%        \centering
%        \includegraphics[width=\linewidth]{figures/ours_vs_default_cd.pdf}
%        \caption{best variant (diagonal, $r$ = 4) against default setting ($r$ = 8) baselines.}
%        \label{fig:cd:ours_vs_default}
%    \end{subfigure}
%    \hfill
%    \begin{subfigure}[b]{0.49\linewidth}
%        \centering
%        \includegraphics[width=\linewidth]{figures/ours_vs_default_cd_r=1.pdf}
%        \caption{LoRA rank = 1}
%        \label{fig:cd:lora1}
%    \end{subfigure}
   \begin{subfigure}[b]{0.49\linewidth}
       \centering
       \includegraphics[width=\linewidth]{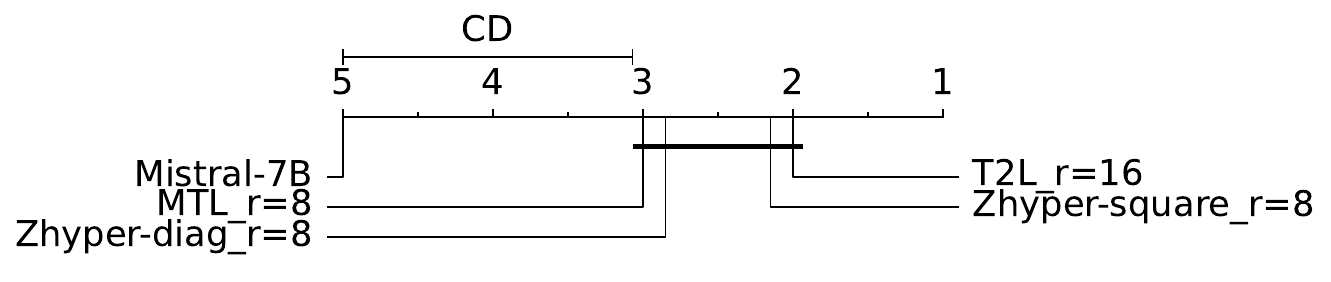}
       \caption{Our method vs. best variants of MTL and T2L.}
       \label{fig:cd:lora4}
   \end{subfigure}
   \hfill
    \begin{subfigure}[b]{0.49\linewidth}
        \centering
        \includegraphics[width=\linewidth]{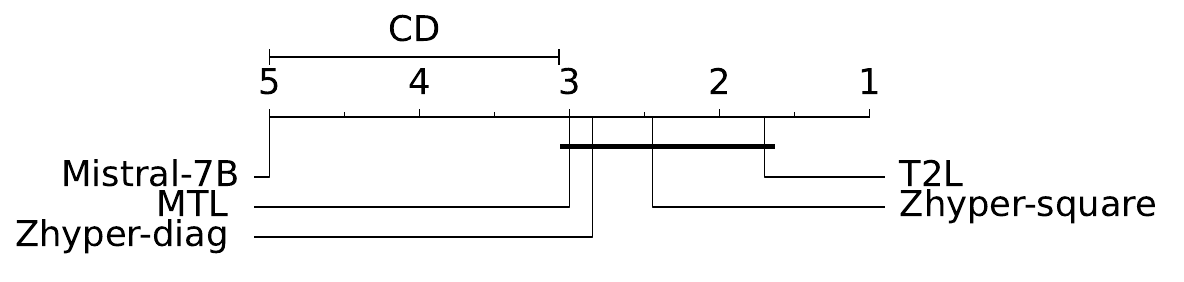}
        \caption{Rank = 8}
        \label{fig:cd:lora8}
    \end{subfigure}\par\bigskip

    \begin{subfigure}[b]{0.49\linewidth}
        \centering
        \includegraphics[width=\linewidth]{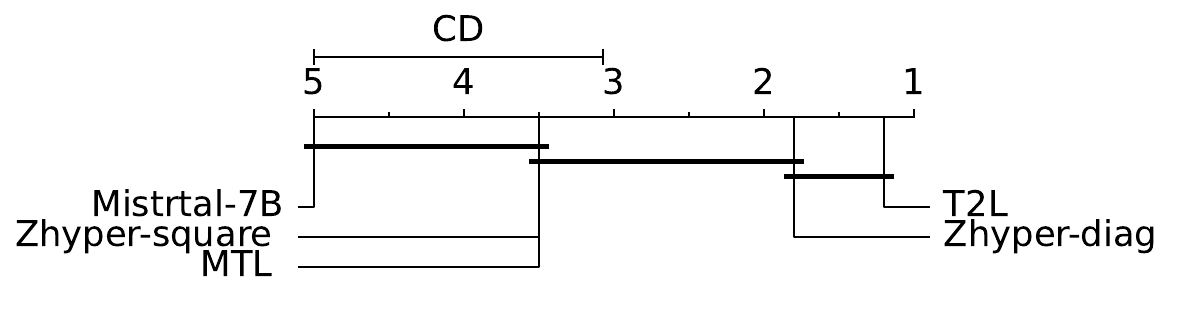}
        \caption{Rank = 16}
        \label{fig:cd:lora16}
    \end{subfigure}
    \begin{subfigure}[b]{0.49\linewidth}
        \centering
        \includegraphics[width=\linewidth]{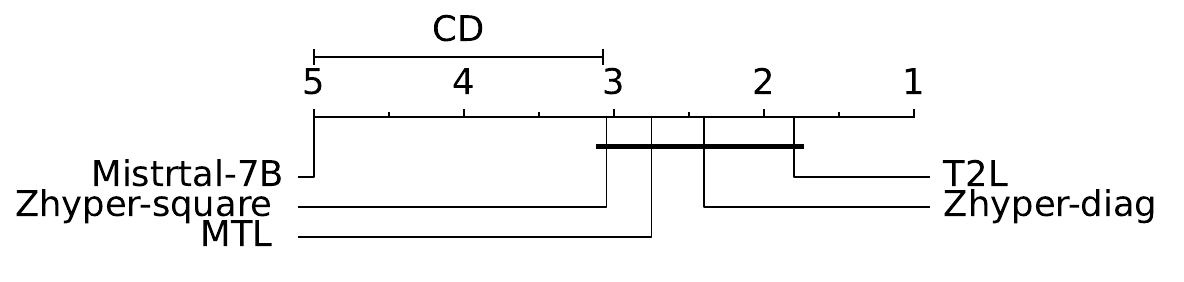}
        \caption{Rank = 32}
        \label{fig:cd:lora32}
    \end{subfigure}
    \caption{Critical Difference (CD) diagrams comparing our method with T2L across LoRA ranks. Lower rank is better. Unconditioned is the base model without any fine-tuning. Groups that are not significantly different are connected by a black bar. }
    \label{fig:cd:ours_vs_baseline_ranks}
\end{figure}

We compare our method against the best-performing T2L model, T2L (SFT) L with $r=16$, which has 110 million trainable parameters. While our method does not fully match T2L’s performance, it achieves comparable results while using 26x fewer trainable parameters and losing only 0.5\% in the average benchmark performance (cf. Table~\ref{tab:t2lvsours}); a full comparison across LoRA ranks is provided in Appendix Section~\ref{subsec:full_task_benchmark}. To assess the significance of this difference, we apply the Friedman test followed by the post hoc Nemenyi test and visualize the results using Critical Difference (CD) diagrams. Black bars connecting different models indicate that there are no statistically significant difference w.r.t. the rank. Our analysis shows that there is no significant difference between our method, T2L, and MTL. Moreover, across LoRA ranks ($r$) 8, 16, and 32, at least one variant of our method is statistically indifferent from T2L as shown in Figure~\ref{fig:cd:ours_vs_baseline_ranks}. Figure~\ref{fig:ours_vs_baseline_scatter} shows that our method is on par with T2L in terms of average benchmark performance, while achieving a high parameter efficiency. The exact number of parameters for each method is listed in Table~\ref{tab:num_params_exact_comaprison}. We note that Hyperdecoders perform strongly; however, they generate a separate LoRA adapter for each problem instance, which is computationally expensive and contrasts with our approach, which generates an adapter from a text description rather than from individual problem instances. Overall, from the results of Tables~\ref{tab:t2lvsours}-\ref{tab:num_params_exact_comaprison} and Figure~\ref{fig:ours_vs_baseline_scatter}, we deduce that our method \ourmethod~offers the best trade-off between parameter-efficiency and accuracy among all considered baselines. 
% \todo{SE: Actually, MTL is the best: it is on par with our method, and uses fewer parameters?}

% \begin{figure} \centering \includegraphics[width=0.50\linewidth]{figures/ours_vs_default_cd.pdf} \caption{TODO (names not fixed)} \label{fig:cd:ours_vs_default} \end{figure}

\subsection{Use case on Cultural Alignment}
\label{subsubsec:exp_hypotheses_align}
%\begin{hypothesis}
%Our conditioning method \todo{SE: our conditioning method equals Zhyper?}
%yields a computationally-efficient strategy for value alignment in LLMs %\textcolor{red}{(Table1)}.
%\end{hypothesis}
%\todo{SE: why is Table 1 in red? It should be Table 3, right?}

%\todo[inline]{todo}

\begin{figure}
    \centering \includegraphics[width=0.8\linewidth]{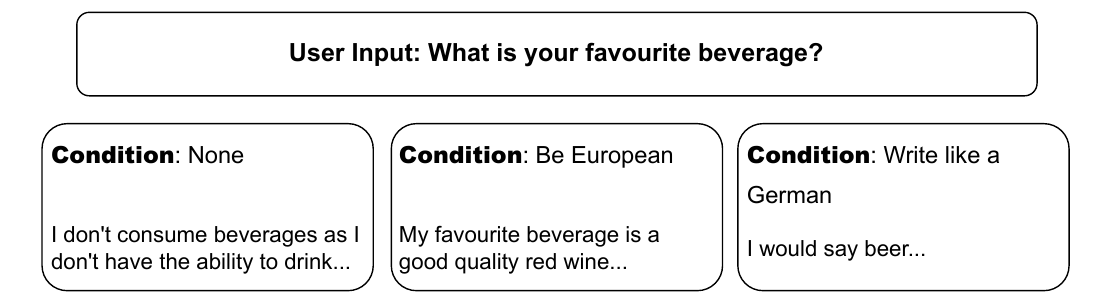}
    \caption{Model output based on text conditions. From left to right: unconditioned model, Europe-conditioned model and Germany-conditioned model. }
    \label{fig:example_chat}
\end{figure}

\textbullet~\emph{Baselines.} Similar to Section~\ref{subsubsec:exp_hypotheses_task}, we use \texttt{\texttt{Mistral-7B-Instruct-v0.2}} as our backbone and unconditional baseline. We include: \textit{Zero-shot}, \textit{Role-play} (prepend a short role specification to the query), \textit{Prepending
culture descriptions}, \textit{Multi-cultural} (MTL), a single LoRA trained on either all countries or all regions, country/region-based oracle, and T2L \citep{charakorn_2025_text2lora}. Additionally, we evaluate a one-hot encoding (OHE) variant of \ourmethod, where the hypernetwork is conditioned on an OHE vector representing the culture.

\textbullet~\emph{Datasets.} We compile a dataset from Reddit's AskX subreddits\footnote{We use Watchful1's reddit dump, which includes data between 2005-06 to 2024-12 for the top 20k subreddits (\url{https://www.reddit.com/r/pushshift/comments/1i4mlqu/dump_files_from_200506_to_202412/})}. We consider the subreddits: r/AskAGerman, r/askmexico, r/AskArgentina, r/AskTurkey, r/AskFrance, r/askegypt, r/AskAJapanese, r/AskIndia, r/AskAChinese, r/AskSouthAfrica, r/askitaly, r/AskARussian, r/AskUK, r/AskAnAmercian, r/asklatinamerica, r/AskAnAfrican, r/AskMiddleEast, r/AskEurope, r/askasia, covering 14 countries and 5 regions/continents. These subreddits were selected based on data availability. We treat each submission title and its top comment as a question-answer pair, considering the top 20k submissions and their top 3 comments based on comment score. To ensure high-quality data, we remove pairs with deleted or removed submissions or comments, as well as pairs containing references to other websites, subreddits, comments, or any type of media, following a filtering procedure similar to OpnionGPT \citep{haller-etal-2024-opiniongpt}. Finally, we randomly select the top 30k pairs per subreddit based on the comment score. To generate cultural descriptions, we prompt \texttt{gpt-4.1-mini} using random pairs sampled from the training dataset. Additionally, we infuse the descriptions with command-like instructions (e.g., \textit{“Write like a German”}), so that the textual conditions reflect both stereotypical cultural traits and explicit commands to emulate the culture. We show examples and the generation prompt in Appendix Section~\ref{subsec:cond_gen}. 
%We evaluate our model on the benchmark dataset CulturalBench \citep{chiu_2025_culturalbench}.

\textbullet~\emph{Evaluation Protocol.} We evaluate cultural alignment on CulturalBench \citep{chiu_2025_culturalbench}, which comprises human-written and human-verified questions spanning 45 regions and 17 topics. The benchmark provides two evaluation setups that share the same underlying questions but differ in querying format: \emph{Easy} uses the original four-way multiple-choice questions, whereas \emph{Hard} converts each question into four binary (True/False) statements, yielding a more challenging setting that reduces shortcutting via option heuristics. 
We report results at both the country and region levels\footnote{In this paper, \emph{country} refers to ISO~3166-1 including administrative countries and territories, whereas \emph{region} denotes macro-regions (e.g., North America, Middle East)}. Accordingly, we train two \ourmethod~models: one on country-level AskX data and one on region-level AskX data. For evaluation, CulturalBench questions are split into \textit{seen} countries/regions (present in training via AskX) and \textit{unseen} countries/regions (absent during training). For text-conditioned models (T2L and Zhyper), we use 12 cultural conditions (see Appendix Section~\ref{subsec:cond_gen} for details) to generate LoRAs per culture (country or region) and report the average performance.
%over the following text conditions: three command-like instructions, four newly generated general descriptions, and four newly generated descriptions appended with command-like instructions. Importantly, the newly generated descriptions are designed to remain general to minimize misalignment between benchmark questions and the conditioning text (see \textcolor{red}{Appendix~Y} for details).

%to assess generalization and the robustness of our model’s cultural alignment across different locales.

%\todo{SE: Caption of Fig. 2 is confusing. Also, according to Ranks, T2L looks better}
%\todo{SE: Table 4 isn't referenced}
%\todo[color=cf]{CF: Table 3 is also not referenced}
%\todo{SE: I don't understand Fig. 3 and what a Critical Difference diagram is. That should be better elaborated I think. Also, which conclusion is there to be drawn?}

\begin{table}[t]
\centering
\caption{\textbf{Cross-country Generalization Results on CulturalBench.} We evaluate Easy/Hard settings and report accuracy (\%) along with the performance rank; Cross-country generalization is assessed by partitioning countries into \emph{seen} and \emph{unseen} groups. Best numbers per column are in \textbf{bold}; second best are \underline{underlined}; values in brackets are mean rank across countries.  “N/A” indicates the setting is not applicable. All methods use a LoRA rank of $r = 8$, unless stated otherwise. Compared to prompt-based baselines and other fine-tuning baselines, \ourmethod~achieves the top scores on all splits and the best averages.} 
\renewcommand{\arraystretch}{1.2}
\resizebox{0.8\textwidth}{!}{%
\begin{tabular}{lcccccc}
\toprule
 & \multicolumn{2}{c}{\textbf{Seen Countries}} & \multicolumn{2}{c}{\textbf{Unseen Countries}} & \multirow{2}{*}{\textbf{Avg. Easy}} & \multirow{2}{*}{\textbf{Avg. Hard}} \\
\cmidrule(lr){2-3} \cmidrule(lr){4-5}
 & \textbf{Easy} & \textbf{Hard} & \textbf{Easy} & \textbf{Hard} &  &  \\
\midrule
Zero-shot & 58.64(6.82) & 34.95(4.96) & 53.93(5.29) & 30.48(3.76) & 55.91(5.77) & 32.36(4.13) \\
Role-play & 64.27(5.54) & 32.81(5.29) & 63.90(3.52) & 29.63(4.03) & 64.06(4.14) & 30.97(4.42) \\
Prepending culture desc. & 64.06(5.54) & 34.07(5.04) & 62.92(3.45) & 31.46(3.92) & 63.40(4.10)  & 32.56(4.27) \\
%3-shot ICL & - & - & - & - & - & - \\
%Multi-cultural(rank=4)& 66.60 & 36.12 & \textbf{65.03} & 33.15 & 65.69 & 34.39 \\
% Multi-cultural (MTL, rank=8) & 66.60 & 36.89 & 64.61 & \underline{34.69} & \underline{65.44} & \underline{35.62} \\
Multi-cultural (MTL, $r = 16$) & 66.21(4.57) & 28.34(6.50) & \underline{64.89}(3.18) & 29.07(4.07) & \underline{65.44}(3.61) & 28.77(4.82) \\
%Culture-specific(Oracle, rank=4) & 68.54 & 29.32 & - & - & - & - \\
T2L & 65.92(4.39) & 34.35(4.61) & 63.58(3.36) & \underline{32.12}(3.23) & 64.56(3.68) & 33.05(3.66) \\
Culture-specific & 67.57(4.04) & 31.84(5.25) & N/A & N/A & N/A & N/A \\

%\ourmethod~(Ours, rank=4) & \textbf{70.63} & \textbf{41.08} & \underline{67.88} & \textbf{36.04} & \textbf{69.03} & \textbf{38.16} \\
\emph{\ourmethod-OHE (Ours)}
& \textbf{70.29}(2.36) & \textbf{40.58}(2.04) & N/A & N/A & 
N/A & N/A \\
\emph{\ourmethod~(Ours)} & \underline{70.15}(2.75) & \underline{40.39}(2.32) & \textbf{67.79}(2.21) & \textbf{36.27}(2.00) & 
\textbf{68.78}(2.38) & \textbf{38.00}(2.10) \\
\bottomrule
\end{tabular}
}
\label{tab:culturalbench-country} 
\end{table}

\textbullet~\emph{Cultural alignment across seen/unseen countries.}
As shown in Table~\ref{tab:culturalbench-country}, our method surpasses prompt-based approaches and fine-tuning baselines across all splits and also leads the averages. Beyond strong results on seen countries, \ourmethod~retains the multi-cultural compatibility that the OHE variant exhibits on seen countries, by conditioning on text, further improving transfer to unseen countries. Notably, the advantage also holds on the Hard split, indicating that the model aligns with cultural norms in a way that remains stable under stricter evaluation rather than relying on surface cues. we show an example generation in Figure~\ref{fig:example_chat}.

\textbullet~\emph{Cultural alignment across seen/unseen regions.}
Table~\ref{tab:culturalbench-region} shows that \ourmethod~attains the best overall average at the regional level and provides balanced improvements over both seen and unseen regions, outperforming prompt-based and other fine-tuning baselines. Crucially, the margin persists on the Hard split, indicating stable regional-level gains under stricter evaluation and complementing the country-level findings under a different partition. An exception is Oceania, where competing MTL variants take the top performance and narrow our margin. We hypothesize that this weaker outcome reflects higher cross-regional transfer difficulty correlated with cultural divergence between Oceania and the training regions.

In both settings, the \ourmethod-OHE variant outperforms Zhyper on seen cultures. However, this method fail to generalize on unseen settings due to the nature of OHE. 

%\paragraph{Cultural alignment}
%Our method demonstrates stronger cultural alignment, particularly on CulturalBench’s \textit{Hard} split, where it attains the best performance across both seen and unseen countries. Because the Hard split suppresses shortcut strategies and emphasizes fine-grained, context-dependent cultural judgments, these gains indicate that our model aligns more precisely with cultural norms rather than relying on surface cues.

%\paragraph{Cross-country generalization (OOD).}
%Although cultural alignment is performed on data from only a subset of countries (in-domain), the model transfers effectively to unseen locales (out-of-domain). The out-of-domain results remain strong and competitive even when the cultural context shifts, suggesting that country-specific alignment induces representations that capture broader cultural regularities and generalize beyond the training countries.

\begin{table}[t]
\centering
\caption{\textbf{Cross-region generalization on CulturalBench.} We evaluate the Easy/Hard settings and report accuracy (\%). Each cell is shown as Easy/Hard. Best numbers per column are in \textbf{bold}; second best are \underline{underlined}. “N/A” indicates the setting is not applicable. All methods use a LoRA rank of $r = 8$, unless stated otherwise. Compared to prompt-based and other fine-tuning baselines, \ourmethod~shows a clear advantage on seen regions and on North America, and achieves the best overall averages.}
\resizebox{\columnwidth}{!}{%
\renewcommand{\arraystretch}{1.2}
\begin{tabular}{lcccccccc}
\toprule
 & \multicolumn{5}{c}{\textbf{Seen Regions}} & \multicolumn{2}{c}{\textbf{Unseen Regions}} & \multirow{2}{*}{\textbf{Avg.}} \\
\cmidrule(lr){2-6} \cmidrule(lr){7-8}
& Latin America & Europe & Africa & Middle East & Asia & N. America & Oceania & \\
\midrule
Zero-shot
& 47.52/20.79 & 56.10/31.01 & 69.40/39.55 & 45.67/21.26 & 54.41/35.08 & 67.11/40.79 & 61.54/\underline{34.62} & 55.91/32.36 \\
Role-play
& 57.43/31.68 & 66.20/32.05 & 73.13/30.60 & 59.84/25.20 & 62.18/29.41 & 64.47/\underline{50.00} & \underline{73.08}/19.23 & 64.06/30.97 \\
Prepending culture desc.
& 59.98/26.57 & 65.24/33.54 & 72.70/31.53 & 57.35/24.54 & 61.55/33.26 & 66.89/48.46 & 70.83/27.24 & 63.60/32.50 \\
%Multi-cultural (rank=4)
%& 59.41/\underline{37.62} & 64.81/\underline{39.72} & \underline{73.13}/36.57 & 55.91/25.98 & 67.02/34.87 & 67.10/44.73 & 76.92/34.62 & 65.60/36.10 \\
% Multi-cultural (MTL, rank=8)
% & 58.42/26.73 & 65.85/32.06 & \underline{77.61}/35.82 & 55.91/22.83 & 65.97/31.51 & \underline{71.05}/\underline{50.00} & \textbf{80.77}/30.77 & \underline{66.18}/31.95 \\
Multi-cultural (MTL, $r = 16$)
& 61.39/32.67 & 65.16/36.24 & 74.63/35.07 & 56.69/27.56 & 67.02/34.24 & \underline{68.42}/46.05 & \textbf{76.92}/\textbf{38.46} & \underline{66.18}/\underline{34.80} \\
T2L
& 60.48/18.98 & 63.73/27.67 & 70.52/19.53 & 57.22/19.03 & 64.90/28.05 & 65.79/29.82 & 65.38/25.00 & 64.15/25.39 \\
Culture-specific 
& \underline{62.38}/\textbf{43.56} & \underline{67.60}/\textbf{41.11} & \underline{77.61}/\underline{39.55} & \underline{61.42}/\textbf{33.86} & 68.07/\textbf{38.03} & N/A & N/A & N/A \\
\emph{\ourmethod-OHE (Ours)} 
& 61.39/\underline{41.58} & \textbf{68.64}/\underline{40.77} & 75.37/\textbf{41.04} & \underline{61.42}/\underline{33.07} & \textbf{69.33}/\underline{37.18} & N/A & N/A & N/A \\
\emph{\ourmethod~(Ours)} 
& \textbf{62.62}/35.97 & \underline{68.23}/38.78 & \textbf{78.05}/38.93 & \textbf{62.14}/29.99 & \underline{68.79}/36.40 & \textbf{71.82}/\textbf{53.40} & 69.23/33.65 & \textbf{68.67}/\textbf{37.52} \\
\bottomrule
\end{tabular}
}
\label{tab:culturalbench-region}
\end{table}

\section{Related Work}
\label{sec:relatedWork}

%\todo[inline, color=cf]{CF: if the eval on survey is no longer in the scope of this paper, which refs do we need or can get rid of?}
\textbf{Low-Rank Adaptation.} To fine-tune LLMs on out-of-distribution applications, %the authors of 
\cite{hu_2021_lora} introduce the concept of Low-Rank Adaptation of LLMs, where the pre-trained LLM  weights are frozen and trainable rank decomposition matrices are introduced. The key concept of LoRA lies in decomposing a weight change matrix $\Delta \mW$ into two low-rank matrices %, referred to as 
$\mA$ and $\mB$.  In \cite{agiza2024mtlora}, the authors extend the LoRA to the multi-task setting by learning shared and task-specific low-rank adapters.

\textbf{Hypernetworks.}
A recent stream of research leverages Hypernetworks that build on the idea of a network's parameters being learned through another neural network~\citep{ha2016hypernetworks}.
%to neural networks whose output defines the parameters of another network
In Text-2-LoRA, \cite{charakorn_2025_text2lora} propose a framework that performs instant adaptation of LLMs from descriptions of downstream tasks. The framework leverages hypernetworks to compress task-specific adapters and enables the zero-shot generation of new LoRA adapters at inference. Hyperdecoders are proposed in \cite{ivison_2022_hyperdecoders} and generate task- and instance-specific decoders showing improved performance in multi-task NLP. Lastly, HyperLoRA leverages hypernetworks for generating task-specific LoRA adapters under low-rank constraints that enable efficient parameter sharing and better cross-task generalization~\citep{lv_2024_hyperlora}.

\begin{table}[t]
    \centering
    \caption{Delinearing our method \ourmethod~from prior works leveraging hypernetworks (Hyperdecoder, HyperLoRA, T2L), and MTLoRA as a multi-task learning approach}
    \resizebox{1\textwidth}{!}{
    \begin{tabular}{lcccc}
    \toprule
         \thead{Model} & \thead{\makecell{Per-context \\materialization}} & \thead{\makecell{Adaptation \\granularity}} & \thead{\makecell{Compact \\ Modulation}} & %\thead{\makecell{Hypernetwork outputs}}\\
         \thead{\makecell{Inference \\ Complexity}}\\
         \midrule
         Hyperdecoder \citep{ivison_2022_hyperdecoders} & very high & per instance & \xmark & High \\
         HyperLoRA\citep{lv_2024_hyperlora} & medium & per-context & \xmark & Medium \\
         MTLoRA \citep{agiza2024mtlora} & low & shared across multiple tasks & \xmark & Low\\
         T2L \citep{charakorn_2025_text2lora} & high & per-context & \xmark & Medium/High \\
         \ourmethod~(Ours) & very low & per-context & \cmark & Low \\
         \bottomrule
    \end{tabular}
}
    \label{tab:hypernetworkComparison}
\end{table}

\textbf{Discussion.} \Cref{tab:hypernetworkComparison} compares Hyperdecoder, HyperLoRA, MTLoRA, T2L, and our method \ourmethod~along the dimension per-context materialization, adaptation granularity, compact modulation, and inference complexity. The key distinction is that prior methods require a hypernetwork to produce full LoRA matrices, while \ourmethod~introduces a compact modulation mechanism, i.e., producing or learning full LoRA matrices vs. \ourmethod's approach of outputting only a compact matrix combining the fixed adapters' weights. Therefore, our framework reduces the number of parameters by an order of magnitude while maintaining competitive accuracy as shown in \Cref{sec:evaluation}.

\textbf{Cultural Alignment of LLMs.}
Evaluations typically use probability surveys \citep{haerpfer_2024_wvs,pewglobalattitudes, durmus2023towards} or non-survey suites built from authored/mined culture questions \citep{pistilli2024civics,ju2025benchmarking,myung2024blend,rao-etal-2025-normad,li2024culturepark}. Surveys are representative but non-everyday questions, focusing on opinions and attitudes, are sensitive to evaluation design\citep{Khan_2025_Randomness}, while many non-survey suites lack rigorous validation. We adopt CulturalBench\citep{chiu_2025_culturalbench} as cultural alignment benchmark for its breadth across countries, regions, and topics and its systematic human–AI red-teaming with a challenging Easy/Hard split.

Methodologically, prior work spans anthropological/persona prompting \citep{alkhamissi-etal-2024-investigating}, survey- or simulation-driven data curation \citep{li2024culturellm,li2024culturepark}, and distributional alignment via self-curated supervision or modified objectives \citep{xu2025self,Yao_2025_CAReDiO,suh-etal-2025-language,cao-etal-2025-specializing}. Our approach instead uses a hypernetwork to generate LoRA adapters from natural-language cultural descriptions at inference time, enabling parameter-efficient per-locale specialization with improved cross-locale generalization.

\section{Conclusion}
\label{sec:conclusion}
Despite the broad success of LLMs, current approaches face persistent challenges in efficiently conditioning LLMs, particularly for content alignment with a large contextual corpus.
We introduce a parameter-efficient factorized hypernetwork framework, called Zhyper, for context-aware LoRA adapters given textual descriptions.
Specifically, we leverage a hypernetwork that yields for each textual description a layer- and target module-specific embedding vector that is injected in LoRA adapters. 
Our evaluation highlights that \ourmethod's computational demands are at an order of magnitude lower -- up to 26x fewer parameters -- compared to existing models while achieving competitive predictive performance. 
Through comprehensive empirical evaluation on task conditioning on 10 benchmark datasets, our method shows competitive results with state-of-the-art, while on a cultural alignment setting, Zhyper shows better generalization capabilities to out-of-domain and unseen contexts. 
%Our findings illustrate the potential of hypernetwork-conditioned LoRA adapters to facilitate dynamic and fine-grained adaptation of LLMs without incurring prohibitive computational costs, and thereby, provide an environmentally friendly solution.
These results highlight the potential of hypernetwork-conditioned LoRA adapters for dynamic, fine-grained LLM adaptation at minimal computational cost, supporting more sustainable and flexible model deployment.

\section{Ethics Statement}
While our method demonstrates improved cultural alignment, we acknowledge that using Reddit as a data source introduces potential biases. We do not filter the dataset for political correctness or linguistic accuracy, therefore some QA pairs may contain harmful content. Although we select the top-voted comments, these can still be conflicting due to the diversity of users’ opinions. Moreover, by relying on Reddit, we model a specific subset of people—those who use the platform—which may not accurately reflect the broader cultural perspectives of the general population.

%\newpage
\bibliographystyle{iclr2026_conference}
\bibliography{iclr2026_conference} 

\begin{thebibliography}{53}
\providecommand{\natexlab}[1]{#1}
\providecommand{\url}[1]{\texttt{#1}}
\expandafter\ifx\csname urlstyle\endcsname\relax
  \providecommand{\doi}[1]{doi: #1}\else
  \providecommand{\doi}{doi: \begingroup \urlstyle{rm}\Url}\fi

\bibitem[Agiza et~al.(2024)Agiza, Neseem, and Reda]{agiza2024mtlora}
Ahmed Agiza, Marina Neseem, and Sherief Reda.
\newblock Mtlora: Low-rank adaptation approach for efficient multi-task learning.
\newblock In \emph{Proceedings of the IEEE/CVF Conference on Computer Vision and Pattern Recognition}, pp.\  16196--16205, 2024.

\bibitem[Ahn et~al.(2024)Ahn, Verma, Lou, Liu, Zhang, and Yin]{ahn-etal-2024-large}
Janice Ahn, Rishu Verma, Renze Lou, Di~Liu, Rui Zhang, and Wenpeng Yin.
\newblock Large language models for mathematical reasoning: Progresses and challenges.
\newblock In Neele Falk, Sara Papi, and Mike Zhang (eds.), \emph{Proceedings of the 18th Conference of the European Chapter of the Association for Computational Linguistics: Student Research Workshop}, pp.\  225--237, St. Julian{'}s, Malta, March 2024. Association for Computational Linguistics.
\newblock \doi{10.18653/v1/2024.eacl-srw.17}.

\bibitem[Alayrac et~al.(2022)Alayrac, Donahue, Luc, Miech, Barr, Hasson, Lenc, Mensch, Millican, Reynolds, Ring, Rutherford, Cabi, Han, Gong, Samangooei, Monteiro, Menick, Borgeaud, Brock, Nematzadeh, Sharifzadeh, Binkowski, Barreira, Vinyals, Zisserman, and Simonyan]{alayrac2022flamingo}
Jean-Baptiste Alayrac, Jeff Donahue, Pauline Luc, Antoine Miech, Iain Barr, Yana Hasson, Karel Lenc, Arthur Mensch, Katie Millican, Malcolm Reynolds, Roman Ring, Eliza Rutherford, Serkan Cabi, Tengda Han, Zhitao Gong, Sina Samangooei, Marianne Monteiro, Jacob Menick, Sebastian Borgeaud, Andrew Brock, Aida Nematzadeh, Sahand Sharifzadeh, Mikolaj Binkowski, Ricardo Barreira, Oriol Vinyals, Andrew Zisserman, and Karen Simonyan.
\newblock Flamingo: a visual language model for few-shot learning.
\newblock In \emph{Advances in Neural Information Processing Systems (NeurIPS)}, 2022.

\bibitem[AlKhamissi et~al.(2024)AlKhamissi, ElNokrashy, Alkhamissi, and Diab]{alkhamissi-etal-2024-investigating}
Badr AlKhamissi, Muhammad ElNokrashy, Mai Alkhamissi, and Mona Diab.
\newblock Investigating cultural alignment of large language models.
\newblock In Lun-Wei Ku, Andre Martins, and Vivek Srikumar (eds.), \emph{Proceedings of the 62nd Annual Meeting of the Association for Computational Linguistics (Volume 1: Long Papers)}, pp.\  12404--12422, Bangkok, Thailand, August 2024. Association for Computational Linguistics.
\newblock \doi{10.18653/v1/2024.acl-long.671}.

\bibitem[Austin et~al.(2021)Austin, Odena, Nye, Bosma, Michalewski, Dohan, Jiang, Cai, Terry, Le, and Sutton]{austin_2021_mbpp}
Jacob Austin, Augustus Odena, Maxwell Nye, Maarten Bosma, Henryk Michalewski, David Dohan, Ellen Jiang, Carrie Cai, Michael Terry, Quoc Le, and Charles Sutton.
\newblock Program synthesis with large language models, 2021.

\bibitem[Bisk et~al.(2019)Bisk, Zellers, Bras, Gao, and Choi]{Bisk_2019_PIQA}
Yonatan Bisk, Rowan Zellers, Ronan~Le Bras, Jianfeng Gao, and Yejin Choi.
\newblock Piqa: Reasoning about physical commonsense in natural language.
\newblock In \emph{AAAI Conference on Artificial Intelligence}, 2019.

\bibitem[Brown et~al.(2020)Brown, Mann, Ryder, Subbiah, Kaplan, Dhariwal, Neelakantan, Shyam, Sastry, Askell, Agarwal, Herbert-Voss, Krueger, Henighan, Child, Ramesh, Ziegler, Wu, Winter, Hesse, Chen, Sigler, Litwin, Gray, Chess, Clark, Berner, McCandlish, Radford, Sutskever, and Amodei]{brown_2020_languagemodelsfewshotlearners}
Tom~B. Brown, Benjamin Mann, Nick Ryder, Melanie Subbiah, Jared Kaplan, Prafulla Dhariwal, Arvind Neelakantan, Pranav Shyam, Girish Sastry, Amanda Askell, Sandhini Agarwal, Ariel Herbert-Voss, Gretchen Krueger, Tom Henighan, Rewon Child, Aditya Ramesh, Daniel~M. Ziegler, Jeffrey Wu, Clemens Winter, Christopher Hesse, Mark Chen, Eric Sigler, Mateusz Litwin, Scott Gray, Benjamin Chess, Jack Clark, Christopher Berner, Sam McCandlish, Alec Radford, Ilya Sutskever, and Dario Amodei.
\newblock Language models are few-shot learners, 2020.

\bibitem[Cao et~al.(2025)Cao, Liu, Arora, Augenstein, R{\"o}ttger, and Hershcovich]{cao-etal-2025-specializing}
Yong Cao, Haijiang Liu, Arnav Arora, Isabelle Augenstein, Paul R{\"o}ttger, and Daniel Hershcovich.
\newblock Specializing large language models to simulate survey response distributions for global populations.
\newblock In Luis Chiruzzo, Alan Ritter, and Lu~Wang (eds.), \emph{Proceedings of the 2025 Conference of the Nations of the Americas Chapter of the Association for Computational Linguistics: Human Language Technologies (Volume 1: Long Papers)}, pp.\  3141--3154, Albuquerque, New Mexico, April 2025. Association for Computational Linguistics.
\newblock ISBN 979-8-89176-189-6.
\newblock \doi{10.18653/v1/2025.naacl-long.162}.

\bibitem[Charakorn et~al.(2025)Charakorn, Cetin, Tang, and Lange]{charakorn_2025_text2lora}
Rujikorn Charakorn, Edoardo Cetin, Yujin Tang, and Robert~Tjarko Lange.
\newblock Text-to-lo{RA}: Instant transformer adaption.
\newblock In \emph{Forty-second International Conference on Machine Learning}, 2025.

\bibitem[Chen et~al.(2021)Chen, Tworek, Jun, Yuan, Pond{\'e}, Kaplan, Edwards, Burda, Joseph, Brockman, Ray, Puri, Krueger, Petrov, Khlaaf, Sastry, Mishkin, Chan, Gray, Ryder, Pavlov, Power, Kaiser, Bavarian, Winter, Tillet, Such, Cummings, Plappert, Chantzis, Barnes, Herbert-Voss, Guss, Nichol, Babuschkin, Balaji, Jain, Carr, Leike, Achiam, Misra, Morikawa, Radford, Knight, Brundage, Murati, Mayer, Welinder, McGrew, Amodei, McCandlish, Sutskever, and Zaremba]{Chen_2021_humaneval}
Mark Chen, Jerry Tworek, Heewoo Jun, Qiming Yuan, Henrique Pond{\'e}, Jared Kaplan, Harrison Edwards, Yura Burda, Nicholas Joseph, Greg Brockman, Alex Ray, Raul Puri, Gretchen Krueger, Michael Petrov, Heidy Khlaaf, Girish Sastry, Pamela Mishkin, Brooke Chan, Scott Gray, Nick Ryder, Mikhail Pavlov, Alethea Power, Lukasz Kaiser, Mo~Bavarian, Clemens Winter, Phil Tillet, Felipe~Petroski Such, David~W. Cummings, Matthias Plappert, Fotios Chantzis, Elizabeth Barnes, Ariel Herbert-Voss, William~H. Guss, Alex Nichol, Igor Babuschkin, Suchir Balaji, Shantanu Jain, Andrew Carr, Jan Leike, Josh Achiam, Vedant Misra, Evan Morikawa, Alec Radford, Matthew~M. Knight, Miles Brundage, Mira Murati, Katie Mayer, Peter Welinder, Bob McGrew, Dario Amodei, Sam McCandlish, Ilya Sutskever, and Wojciech Zaremba.
\newblock Evaluating large language models trained on code.
\newblock \emph{ArXiv}, abs/2107.03374, 2021.

\bibitem[Chiu et~al.(2025)Chiu, Jiang, Lin, Park, Li, Ravi, Bhatia, Antoniak, Tsvetkov, Shwartz, and Choi]{chiu_2025_culturalbench}
Yu~Ying Chiu, Liwei Jiang, Bill~Yuchen Lin, Chan~Young Park, Shuyue~Stella Li, Sahithya Ravi, Mehar Bhatia, Maria Antoniak, Yulia Tsvetkov, Vered Shwartz, and Yejin Choi.
\newblock Culturalbench: A robust, diverse, and challenging cultural benchmark by human-ai culturalteaming, 2025.

\bibitem[Clark et~al.(2019)Clark, Lee, Chang, Kwiatkowski, Collins, and Toutanova]{clark_2019_boolq}
Christopher Clark, Kenton Lee, Ming-Wei Chang, Tom Kwiatkowski, Michael Collins, and Kristina Toutanova.
\newblock {B}ool{Q}: Exploring the surprising difficulty of natural yes/no questions.
\newblock In Jill Burstein, Christy Doran, and Thamar Solorio (eds.), \emph{Proceedings of the 2019 Conference of the North {A}merican Chapter of the Association for Computational Linguistics: Human Language Technologies, Volume 1 (Long and Short Papers)}, pp.\  2924--2936, Minneapolis, Minnesota, June 2019. Association for Computational Linguistics.
\newblock \doi{10.18653/v1/N19-1300}.

\bibitem[Clark et~al.(2018)Clark, Cowhey, Etzioni, Khot, Sabharwal, Schoenick, and Tafjord]{Clark_2018_Arc}
Peter Clark, Isaac Cowhey, Oren Etzioni, Tushar Khot, Ashish Sabharwal, Carissa Schoenick, and Oyvind Tafjord.
\newblock Think you have solved question answering? try arc, the ai2 reasoning challenge.
\newblock \emph{ArXiv}, abs/1803.05457, 2018.

\bibitem[Cobbe et~al.(2021)Cobbe, Kosaraju, Bavarian, Chen, Jun, Kaiser, Plappert, Tworek, Hilton, Nakano, Hesse, and Schulman]{Cobbe_2021_TrainingVT}
Karl Cobbe, Vineet Kosaraju, Mo~Bavarian, Mark Chen, Heewoo Jun, Lukasz Kaiser, Matthias Plappert, Jerry Tworek, Jacob Hilton, Reiichiro Nakano, Christopher Hesse, and John Schulman.
\newblock Training verifiers to solve math word problems.
\newblock \emph{ArXiv}, abs/2110.14168, 2021.

\bibitem[Ding et~al.(2023)Ding, Qin, Yang, Wei, Yang, Su, Hu, Chen, Chan, Chen, Yi, Zhao, Wang, Liu, Zheng, Chen, Liu, Tang, Li, and Sun]{Ding_2023_Parameter-efficient}
Ning Ding, Yujia Qin, Guang Yang, Fu~Wei, Zonghan Yang, Yusheng Su, Shengding Hu, Yulin Chen, Chi-Min Chan, Weize Chen, Jing Yi, Weilin Zhao, Xiaozhi Wang, Zhiyuan Liu, Haitao Zheng, Jianfei Chen, Y.~Liu, Jie Tang, Juanzi Li, and Maosong Sun.
\newblock Parameter-efficient fine-tuning of large-scale pre-trained language models.
\newblock \emph{Nature Machine Intelligence}, 5:\penalty0 220--235, 2023.
\newblock \doi{10.1038/s42256-023-00626-4}.

\bibitem[Dong et~al.(2024)Dong, Li, Dai, Zheng, Ma, Li, Xia, Xu, Wu, Chang, Sun, Li, and Sui]{dong_2024_ICLsurvey}
Qingxiu Dong, Lei Li, Damai Dai, Ce~Zheng, Jingyuan Ma, Rui Li, Heming Xia, Jingjing Xu, Zhiyong Wu, Baobao Chang, Xu~Sun, Lei Li, and Zhifang Sui.
\newblock A survey on in-context learning.
\newblock In Yaser Al-Onaizan, Mohit Bansal, and Yun-Nung Chen (eds.), \emph{Proceedings of the 2024 Conference on Empirical Methods in Natural Language Processing}, pp.\  1107--1128, Miami, Florida, USA, November 2024. Association for Computational Linguistics.
\newblock \doi{10.18653/v1/2024.emnlp-main.64}.

\bibitem[Durmus et~al.(2023)Durmus, Nguyen, Liao, Schiefer, Askell, Bakhtin, Chen, Hatfield-Dodds, Hernandez, Joseph, et~al.]{durmus2023towards}
Esin Durmus, Karina Nguyen, Thomas~I Liao, Nicholas Schiefer, Amanda Askell, Anton Bakhtin, Carol Chen, Zac Hatfield-Dodds, Danny Hernandez, Nicholas Joseph, et~al.
\newblock Towards measuring the representation of subjective global opinions in language models.
\newblock \emph{arXiv preprint arXiv:2306.16388}, 2023.

\bibitem[Eger et~al.(2025)Eger, Cao, D'Souza, Geiger, Greisinger, Gross, Hou, Krenn, Lauscher, Li, Lin, Moosavi, Zhao, and Miller]{eger2025transformingsciencelargelanguage}
Steffen Eger, Yong Cao, Jennifer D'Souza, Andreas Geiger, Christian Greisinger, Stephanie Gross, Yufang Hou, Brigitte Krenn, Anne Lauscher, Yizhi Li, Chenghua Lin, Nafise~Sadat Moosavi, Wei Zhao, and Tristan Miller.
\newblock Transforming science with large language models: A survey on ai-assisted scientific discovery, experimentation, content generation, and evaluation, 2025.

\bibitem[Geng et~al.(2025)Geng, Chen, Wu, Wan, Zhou, and Chen]{geng-etal-2025-impact}
Mingmeng Geng, Caixi Chen, Yanru Wu, Yao Wan, Pan Zhou, and Dongping Chen.
\newblock The impact of large language models in academia: from writing to speaking.
\newblock In Wanxiang Che, Joyce Nabende, Ekaterina Shutova, and Mohammad~Taher Pilehvar (eds.), \emph{Findings of the Association for Computational Linguistics: ACL 2025}, pp.\  19303--19319, Vienna, Austria, July 2025. Association for Computational Linguistics.
\newblock ISBN 979-8-89176-256-5.
\newblock \doi{10.18653/v1/2025.findings-acl.987}.

\bibitem[Gu et~al.(2025)Gu, Jiang, Shi, Tan, Zhai, Xu, Li, Shen, Ma, Liu, Wang, Zhang, Wang, Gao, Ni, and Guo]{gu2025surveyllmasajudge}
Jiawei Gu, Xuhui Jiang, Zhichao Shi, Hexiang Tan, Xuehao Zhai, Chengjin Xu, Wei Li, Yinghan Shen, Shengjie Ma, Honghao Liu, Saizhuo Wang, Kun Zhang, Yuanzhuo Wang, Wen Gao, Lionel Ni, and Jian Guo.
\newblock A survey on llm-as-a-judge, 2025.

\bibitem[Gugger et~al.(2022)Gugger, Debut, Wolf, Schmid, Mueller, Mangrulkar, Sun, and Bossan]{accelerate}
Sylvain Gugger, Lysandre Debut, Thomas Wolf, Philipp Schmid, Zachary Mueller, Sourab Mangrulkar, Marc Sun, and Benjamin Bossan.
\newblock Accelerate: Training and inference at scale made simple, efficient and adaptable.
\newblock \url{https://github.com/huggingface/accelerate}, 2022.

\bibitem[Ha et~al.(2016)Ha, Dai, and Le]{ha2016hypernetworks}
David Ha, Andrew Dai, and Quoc~V. Le.
\newblock Hypernetworks, 2016.

\bibitem[Haerpfer et~al.(2024)Haerpfer, Inglehart, Moreno, Welzel, Kizilova, Diez-Medrano, Lagos, Norris, Ponarin, and Puranen]{haerpfer_2024_wvs}
Christian Haerpfer, Ronald Inglehart, Alejandro Moreno, Christian Welzel, Kseniya Kizilova, Jaime Diez-Medrano, Marta Lagos, Pippa Norris, Eduard Ponarin, and Bi~Puranen.
\newblock World values survey wave 7 (2017-2022) cross-national data-set, 2024.

\bibitem[Haller et~al.(2024)Haller, Aynetdinov, and Akbik]{haller-etal-2024-opiniongpt}
Patrick Haller, Ansar Aynetdinov, and Alan Akbik.
\newblock {O}pinion{GPT}: Modelling explicit biases in instruction-tuned {LLM}s.
\newblock In Kai-Wei Chang, Annie Lee, and Nazneen Rajani (eds.), \emph{Proceedings of the 2024 Conference of the North American Chapter of the Association for Computational Linguistics: Human Language Technologies (Volume 3: System Demonstrations)}, pp.\  78--86, Mexico City, Mexico, June 2024. Association for Computational Linguistics.
\newblock \doi{10.18653/v1/2024.naacl-demo.8}.
\newblock URL \url{https://aclanthology.org/2024.naacl-demo.8/}.

\bibitem[Hu et~al.(2021)Hu, Shen, Wallis, Allen-Zhu, Li, Wang, Wang, and Chen]{hu_2021_lora}
Edward~J. Hu, Yelong Shen, Phillip Wallis, Zeyuan Allen-Zhu, Yuanzhi Li, Shean Wang, Lu~Wang, and Weizhu Chen.
\newblock Lora: Low-rank adaptation of large language models, 2021.

\bibitem[Ivison \& Peters(2022)Ivison and Peters]{ivison_2022_hyperdecoders}
Hamish Ivison and Matthew~E. Peters.
\newblock Hyperdecoders: Instance-specific decoders for multi-task nlp, 2022.

\bibitem[Jiang et~al.(2023)Jiang, Sablayrolles, Mensch, Bamford, Chaplot, de~las Casas, Bressand, Lengyel, Lample, Saulnier, Lavaud, Lachaux, Stock, Scao, Lavril, Wang, Lacroix, and Sayed]{jiang_2023_mistral7b}
Albert~Q. Jiang, Alexandre Sablayrolles, Arthur Mensch, Chris Bamford, Devendra~Singh Chaplot, Diego de~las Casas, Florian Bressand, Gianna Lengyel, Guillaume Lample, Lucile Saulnier, Lélio~Renard Lavaud, Marie-Anne Lachaux, Pierre Stock, Teven~Le Scao, Thibaut Lavril, Thomas Wang, Timothée Lacroix, and William~El Sayed.
\newblock Mistral 7b, 2023.

\bibitem[Ju et~al.(2025)Ju, Shi, Liu, Ji, Zhang, Zhang, Xu, Yang, Han, and Guo]{ju2025benchmarking}
Chengyi Ju, Weijie Shi, Chengzhong Liu, Jiaming Ji, Jipeng Zhang, Ruiyuan Zhang, Jiajie Xu, Yaodong Yang, Sirui Han, and Yike Guo.
\newblock Benchmarking multi-national value alignment for large language models.
\newblock In \emph{Findings of the Association for Computational Linguistics: ACL 2025}, pp.\  20042--20058, 2025.

\bibitem[Khan et~al.(2025)Khan, Casper, and Hadfield-Menell]{Khan_2025_Randomness}
Ariba Khan, Stephen Casper, and Dylan Hadfield-Menell.
\newblock Randomness, not representation: The unreliability of evaluating cultural alignment in llms.
\newblock \emph{ArXiv}, abs/2503.08688, 2025.
\newblock \doi{10.48550/arXiv.2503.08688}.

\bibitem[Li et~al.(2024{\natexlab{a}})Li, Chen, Wang, Sitaram, and Xie]{li2024culturellm}
Cheng Li, Mengzhuo Chen, Jindong Wang, Sunayana Sitaram, and Xing Xie.
\newblock Culturellm: Incorporating cultural differences into large language models.
\newblock \emph{Advances in Neural Information Processing Systems}, 37:\penalty0 84799--84838, 2024{\natexlab{a}}.

\bibitem[Li et~al.(2024{\natexlab{b}})Li, Teney, Yang, Wen, Xie, and Wang]{li2024culturepark}
Cheng Li, Damien Teney, Linyi Yang, Qingsong Wen, Xing Xie, and Jindong Wang.
\newblock Culturepark: Boosting cross-cultural understanding in large language models.
\newblock \emph{Advances in Neural Information Processing Systems}, 37:\penalty0 65183--65216, 2024{\natexlab{b}}.

\bibitem[Li et~al.(2023)Li, Zhang, Zhang, Long, Xie, and Zhang]{li2023towards}
Zehan Li, Xin Zhang, Yanzhao Zhang, Dingkun Long, Pengjun Xie, and Meishan Zhang.
\newblock Towards general text embeddings with multi-stage contrastive learning.
\newblock \emph{arXiv preprint arXiv:2308.03281}, 2023.

\bibitem[Luo et~al.(2024)Luo, Lei, Lei, Liu, He, Zhao, and Liu]{luo2024moelora}
Tongxu Luo, Jiahe Lei, Fangyu Lei, Weihao Liu, Shizhu He, Jun Zhao, and Kang Liu.
\newblock Moelora: Contrastive learning guided mixture of experts on parameter-efficient fine-tuning for large language models, 2024.

\bibitem[Lv et~al.(2024)Lv, Li, Zhang, Chen, Qi, Zhang, and Zheng]{lv_2024_hyperlora}
Chuancheng Lv, Lei Li, Shitou Zhang, Gang Chen, Fanchao Qi, Ningyu Zhang, and Hai-Tao Zheng.
\newblock {H}yper{L}o{RA}: Efficient cross-task generalization via constrained low-rank adapters generation.
\newblock In Yaser Al-Onaizan, Mohit Bansal, and Yun-Nung Chen (eds.), \emph{Findings of the Association for Computational Linguistics: EMNLP 2024}, pp.\  16376--16393, Miami, Florida, USA, November 2024. Association for Computational Linguistics.
\newblock \doi{10.18653/v1/2024.findings-emnlp.956}.

\bibitem[Mihaylov et~al.(2018)Mihaylov, Clark, Khot, and Sabharwal]{mihaylov_2018_openbook}
Todor Mihaylov, Peter Clark, Tushar Khot, and Ashish Sabharwal.
\newblock Can a suit of armor conduct electricity? a new dataset for open book question answering.
\newblock In Ellen Riloff, David Chiang, Julia Hockenmaier, and Jun{'}ichi Tsujii (eds.), \emph{Proceedings of the 2018 Conference on Empirical Methods in Natural Language Processing}, pp.\  2381--2391, Brussels, Belgium, October-November 2018. Association for Computational Linguistics.
\newblock \doi{10.18653/v1/D18-1260}.

\bibitem[Myung et~al.(2024)Myung, Lee, Zhou, Jin, Putri, Antypas, Borkakoty, Kim, Perez-Almendros, Ayele, et~al.]{myung2024blend}
Junho Myung, Nayeon Lee, Yi~Zhou, Jiho Jin, Rifki Putri, Dimosthenis Antypas, Hsuvas Borkakoty, Eunsu Kim, Carla Perez-Almendros, Abinew~Ali Ayele, et~al.
\newblock Blend: A benchmark for llms on everyday knowledge in diverse cultures and languages.
\newblock \emph{Advances in Neural Information Processing Systems}, 37:\penalty0 78104--78146, 2024.

\bibitem[Ostapenko et~al.(2024)Ostapenko, Su, Ponti, Charlin, Roux, Pereira, Caccia, and Sordoni]{ostapenko_2024_arrow}
Oleksiy Ostapenko, Zhan Su, Edoardo~Maria Ponti, Laurent Charlin, Nicolas~Le Roux, Matheus Pereira, Lucas Caccia, and Alessandro Sordoni.
\newblock Towards modular llms by building and reusing a library of loras, 2024.

\bibitem[{Pew Research Center}(2024)]{pewglobalattitudes}
{Pew Research Center}.
\newblock Pew research center global attitudes survey: Datasets portal.
\newblock \url{https://www.pewresearch.org/global/datasets/}, 2024.
\newblock Accessed 2025-09-08.

\bibitem[Pistilli et~al.(2024)Pistilli, Leidinger, Jernite, Kasirzadeh, Luccioni, and Mitchell]{pistilli2024civics}
Giada Pistilli, Alina Leidinger, Yacine Jernite, Atoosa Kasirzadeh, Alexandra~Sasha Luccioni, and Margaret Mitchell.
\newblock Civics: Building a dataset for examining culturally-informed values in large language models.
\newblock In \emph{Proceedings of the AAAI/ACM Conference on AI, Ethics, and Society}, volume~7, pp.\  1132--1144, 2024.

\bibitem[Prottasha et~al.(2024)Prottasha, Mahmud, Sobuj, Bhat, Kowsher, Yousefi, and Garibay]{Prottasha2024Parameter-efficient}
Nusrat~Jahan Prottasha, Asif Mahmud, Md. Shohanur~Islam Sobuj, Prakash Bhat, Md. Kowsher, Niloofar Yousefi, and O.~Garibay.
\newblock Parameter-efficient fine-tuning of large language models using semantic knowledge tuning.
\newblock \emph{Scientific Reports}, 14, 2024.
\newblock \doi{10.1038/s41598-024-75599-4}.

\bibitem[Rao et~al.(2025)Rao, Yerukola, Shah, Reinecke, and Sap]{rao-etal-2025-normad}
Abhinav~Sukumar Rao, Akhila Yerukola, Vishwa Shah, Katharina Reinecke, and Maarten Sap.
\newblock {N}orm{A}d: A framework for measuring the cultural adaptability of large language models.
\newblock In Luis Chiruzzo, Alan Ritter, and Lu~Wang (eds.), \emph{Proceedings of the 2025 Conference of the Nations of the Americas Chapter of the Association for Computational Linguistics: Human Language Technologies (Volume 1: Long Papers)}, pp.\  2373--2403, Albuquerque, New Mexico, April 2025. Association for Computational Linguistics.
\newblock ISBN 979-8-89176-189-6.
\newblock \doi{10.18653/v1/2025.naacl-long.120}.

\bibitem[Rozière et~al.(2023)Rozière, Gehring, Gloeckle, Sootla, Gat, Tan, Adi, Liu, Sauvestre, Remez, Rapin, Kozhevnikov, Evtimov, Bitton, Bhatt, Ferrer, Grattafiori, Xiong, Défossez, Copet, Azhar, Touvron, Martin, Usunier, Scialom, and Synnaeve]{roziere2023code}
Baptiste Rozière, Jonas Gehring, Fabian Gloeckle, Sten Sootla, Itai Gat, Xiaoqing~Ellen Tan, Yossi Adi, Jingyu Liu, Romain Sauvestre, Tal Remez, Jérémy Rapin, Artyom Kozhevnikov, Ivan Evtimov, Joanna Bitton, Manish Bhatt, Cristian~Canton Ferrer, Aaron Grattafiori, Wenhan Xiong, Alexandre Défossez, Jade Copet, Faisal Azhar, Hugo Touvron, Louis Martin, Nicolas Usunier, Thomas Scialom, and Gabriel Synnaeve.
\newblock Code llama: Open foundation models for code.
\newblock \emph{arXiv preprint}, arXiv:2308.12950, 2023.

\bibitem[Sakaguchi et~al.(2021)Sakaguchi, Bras, Bhagavatula, and Choi]{keisuke_2021_winogrande}
Keisuke Sakaguchi, Ronan~Le Bras, Chandra Bhagavatula, and Yejin Choi.
\newblock Winogrande: an adversarial winograd schema challenge at scale.
\newblock \emph{Commun. ACM}, 64\penalty0 (9):\penalty0 99–106, August 2021.
\newblock ISSN 0001-0782.
\newblock \doi{10.1145/3474381}.

\bibitem[Shalev-Shwartz \& Ben-David(2014)Shalev-Shwartz and Ben-David]{shwartz_2020_bible}
Shai Shalev-Shwartz and Shai Ben-David.
\newblock \emph{Understanding Machine Learning - From Theory to Algorithms.}
\newblock Cambridge University Press, 2014.
\newblock ISBN 978-1-10-705713-5.

\bibitem[Suh et~al.(2025)Suh, Jahanparast, Moon, Kang, and Chang]{suh-etal-2025-language}
Joseph Suh, Erfan Jahanparast, Suhong Moon, Minwoo Kang, and Serina Chang.
\newblock Language model fine-tuning on scaled survey data for predicting distributions of public opinions.
\newblock In Wanxiang Che, Joyce Nabende, Ekaterina Shutova, and Mohammad~Taher Pilehvar (eds.), \emph{Proceedings of the 63rd Annual Meeting of the Association for Computational Linguistics (Volume 1: Long Papers)}, pp.\  21147--21170, Vienna, Austria, July 2025. Association for Computational Linguistics.
\newblock ISBN 979-8-89176-251-0.
\newblock \doi{10.18653/v1/2025.acl-long.1028}.

\bibitem[Wang et~al.(2024)Wang, Zhao, Wang, Wang, and Liu]{wang2024malora}
Xujia Wang, Haiyan Zhao, Shuo Wang, Hanqing Wang, and Zhiyuan Liu.
\newblock Malora: Mixture of asymmetric low-rank adaptation for enhanced multi-task learning, 2024.

\bibitem[Wang et~al.(2023)Wang, Lin, Zeng, and Zhang]{wang2023multilora}
Yiming Wang, Yu~Lin, Xiaodong Zeng, and Guannan Zhang.
\newblock Multilora: Democratizing lora for better multi-task learning, 2023.

\bibitem[Wang et~al.(2022)Wang, Mishra, Alipoormolabashi, Kordi, Mirzaei, Arunkumar, Ashok, Dhanasekaran, Naik, Stap, et~al.]{Wang_2022_sni}
Yizhong Wang, Swaroop Mishra, Pegah Alipoormolabashi, Yeganeh Kordi, Amirreza Mirzaei, Anjana Arunkumar, Arjun Ashok, Arut~Selvan Dhanasekaran, Atharva Naik, David Stap, et~al.
\newblock Super-naturalinstructions:generalization via declarative instructions on 1600+ tasks.
\newblock In \emph{EMNLP}, 2022.

\bibitem[Weyssow et~al.(2023)Weyssow, Zhou, Kim, Lo, and Sahraoui]{Weyssow2023Exploring}
M.~Weyssow, Xin Zhou, Kisub Kim, David Lo, and H.~Sahraoui.
\newblock Exploring parameter-efficient fine-tuning techniques for code generation with large language models.
\newblock \emph{ACM Transactions on Software Engineering and Methodology}, 2023.
\newblock \doi{10.1145/3714461}.

\bibitem[Xu et~al.(2025)Xu, Leng, Yu, and Xiong]{xu2025self}
Shaoyang Xu, Yongqi Leng, Linhao Yu, and Deyi Xiong.
\newblock Self-pluralising culture alignment for large language models.
\newblock In \emph{Proceedings of the 2025 Conference of the Nations of the Americas Chapter of the Association for Computational Linguistics: Human Language Technologies (Volume 1: Long Papers)}, pp.\  6859--6877, 2025.

\bibitem[Yao et~al.(2025)Yao, Yi, Wang, Dou, and Xie]{Yao_2025_CAReDiO}
Jing Yao, Xiaoyuan Yi, Jindong Wang, Zhicheng Dou, and Xing Xie.
\newblock Caredio: Cultural alignment of llm via representativeness and distinctiveness guided data optimization.
\newblock \emph{ArXiv}, abs/2504.08820, 2025.
\newblock \doi{10.48550/arXiv.2504.08820}.

\bibitem[Zellers et~al.(2019)Zellers, Holtzman, Bisk, Farhadi, and Choi]{Zellers_2019_HellaSwag}
Rowan Zellers, Ari Holtzman, Yonatan Bisk, Ali Farhadi, and Yejin Choi.
\newblock Hellaswag: Can a machine really finish your sentence?
\newblock In \emph{Annual Meeting of the Association for Computational Linguistics}, 2019.

\bibitem[Zhang et~al.(2024)Zhang, Zhang, Long, Xie, Dai, Tang, Lin, Yang, Xie, Huang, et~al.]{zhang2024mgte}
Xin Zhang, Yanzhao Zhang, Dingkun Long, Wen Xie, Ziqi Dai, Jialong Tang, Huan Lin, Baosong Yang, Pengjun Xie, Fei Huang, et~al.
\newblock mgte: Generalized long-context text representation and reranking models for multilingual text retrieval.
\newblock \emph{arXiv preprint arXiv:2407.19669}, 2024.

\end{thebibliography}
\appendix

\section{Training Procedure And Hyperparameter}
We use the following hyperparameters to train our model (Table~\ref{tab:hyper_params}). Notably, we train the model for 2,000 epochs for tasks and 5,000 epochs for cultures. For LoRA ranks below 16, training fits on a single H100 GPU (80 GB VRAM). To accelerate training, we distribute it across 8 H100 GPUs using Accelerate \citep{accelerate}. For example, training with LoRA rank 8 on the tasks dataset takes approximately 7–8 hours of wall-clock time, otherwise on 1 GPU, whereas on a single GPU it can take up to 48 hours.

\begin{table}[!h]
    \centering
\caption{Hyperparameters used during training. 
$d_{\text{MLP\_out}}$ denotes the output dimension of the final MLP block, which serves as input to the network's output head. 
$d_{\text{MLP\_in}}$ denotes the input dimension of each MLP block. 
$d_{\text{MLP\_hidden}}$ denotes the hidden dimension of each MLP block.}
    \resizebox{0.6\textwidth}{!}{%
    \begin{tabular}{c c c}
        \toprule
        \thead{Hyperparameter} & \thead{Ours/T2L} & \thead{Task/Culture-specific} \\
        \midrule
        Max learning rate  & 2.5e-5 & 3e-5   \\
        Gradient accumulation steps & 1 & 1  \\
        Batch size & 8 & 8  \\
        NEFTune noise alpha & 5.0 & 5.0 \\
        Warmup fraction & 0.2 & 0.1 \\
        Label smoothing & 0.1 & 0.1 \\
        Weight decay & 0.1 & 0.1 \\
        $d_{\text{MLP\_out}}$ & 512 & N/A \\
        $d_{\text{MLP\_in}}$ & 128 & N/A \\
        $d_{\text{MLP\_hidden}}$ & 512 & N/A \\
        \bottomrule
    \end{tabular}%
    }
    \label{tab:hyper_params}
\end{table}
\clearpage
\section{Hyperparameter Tuning}
\label{subsec:hyperparam_tuning}
\begin{figure}
    \centering \includegraphics[width=\linewidth]{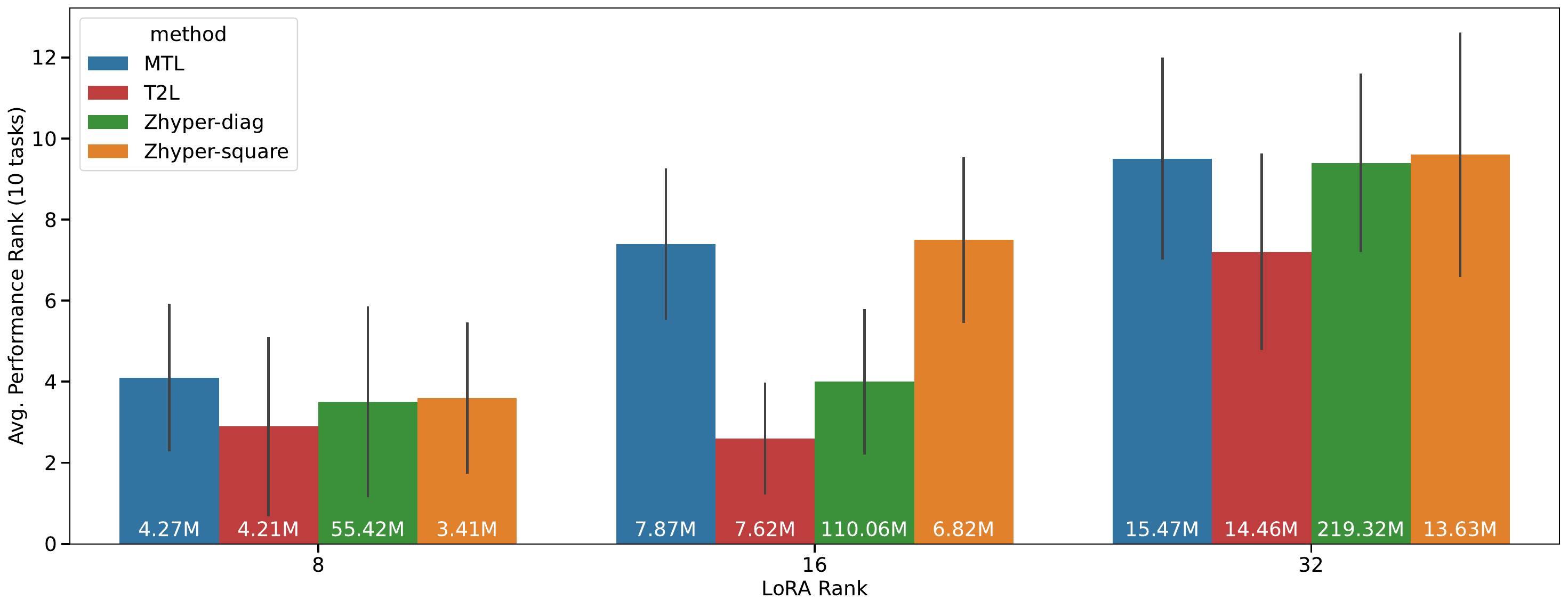}
    \caption{Average Performance Rank on the benchmark validation set (lower is better). For MTL, best variant is at $r=8$, T2L, $r=16$ and Zhyper $r=8, diag$.}
    \label{fig:val_set_task}
\end{figure}

We report the performance of MTL, T2L, and Zhyper on a subset of the benchmark validation set (Figure~\ref{fig:val_set_task}). For Table~\ref{tab:t2lvsours}, we select the best-performing variant of each method. For cultural alignment, since CulturalBench is relatively small, containing up to 200 questions per country, we do not use it as a validation set. Instead, we sample 10\% of the training data (subreddit QA pairs) as a validation set and use SFT loss as the evaluation metric. The best-performing variant of each method is then used in the benchmarking tables. Table~\ref{tab:val_loss_region_country} reports the performance of all methods for both country- and region-based models.

% \begin{figure}[t]
%     \centering
%     \begin{subfigure}[b]{0.495\linewidth}
%         \centering
%         \includegraphics[width=\linewidth]{figures/val_set_methods_region.pdf}
%     \end{subfigure}
%     \begin{subfigure}[b]{0.495\linewidth}
%         \centering
%         \includegraphics[width=\linewidth]{figures/val_set_methods_region.pdf}
%     \end{subfigure}
%     \caption{SFT Loss of the last step for culture alignment models. TODO}
%     \label{fig:val_set_region_country}
% \end{figure}

\begin{table}[t]
  \centering
  \caption{SFT loss by LoRA rank $(r)$ (lower is better). Left: country models; right: region models. \textbf{Bold} indicates the best performance across LoRA ranks for each method. For \ourmethod, the best variant is reported considering both the Z matrix type (diag or square) and the LoRA rank. That is, \ourmethod-diag with LoRA rank 8 achieves the best performance for both country- and region-based models.}
  \begin{minipage}{0.45\textwidth}
    \centering
    \resizebox{\textwidth}{!}{%
    \begin{tabular}{l r r r}
      \toprule
      \multirow{2}{*}{\textbf{Method}} & \multicolumn{3}{c}{\textbf{Rank}} \\
      \cmidrule(lr){2-4}
                                       & \textbf{8} & \textbf{16} & \textbf{32} \\
      \midrule
      MTL           & 2.748 & \bf 2.688 & 2.759 \\
      T2L           & \bf 2.764 & 2.777 & 2.775 \\
      Zhyper-diag   & \bf 2.705 & 2.726 & 2.734 \\
      Zhyper-square & \bf 2.731 & 2.730 & 2.734 \\
      \bottomrule
    \end{tabular}}
  \end{minipage}\hfill
  \begin{minipage}{0.45\textwidth}
    \centering
    \resizebox{\textwidth}{!}{%
    \begin{tabular}{l r r r}
      \toprule
      \multirow{2}{*}{\textbf{Method}} & \multicolumn{3}{c}{\textbf{Rank}} \\
      \cmidrule(lr){2-4}
                                       & \textbf{8} & \textbf{16} & \textbf{32} \\
      \midrule
      MTL           & 2.773 & \bf 2.753 & 2.772 \\
      T2L           & \bf 2.815 & 2.880 & 2.826 \\
      Zhyper-diag   & \bf 2.756 & 2.818 & 2.766 \\
      Zhyper-square & 2.765 & 2.815 & \bf 2.764 \\
      \bottomrule
    \end{tabular}}
  \end{minipage}
    \label{tab:val_loss_region_country}
\end{table}

% \begin{table}[t]
%   \centering
%   \begin{minipage}{0.4\textwidth}
%     \centering
%     \caption{SFT loss by rank}
%     {\small
%     \begin{tabular}{l r r r}
%       \toprule
%       \multirow{2}{*}{\textbf{Method}} & \multicolumn{3}{c}{\textbf{Rank}} \\
%       \cmidrule(lr){2-4}
%                                        & \textbf{8} & \textbf{16} & \textbf{32} \\
%       \midrule
%       MTL           & 2.747905 & 2.688498 & 2.759371 \\
%       T2L           & 2.763771 & 2.776751 & 2.775124 \\
%       Zhyper-diag   & 2.705015 & 2.726074 & 2.733713 \\
%       Zhyper-square & 2.730687 & 2.730323 & 2.734119 \\
%       \bottomrule
%     \end{tabular}}
%     \label{tab:left}
%   \end{minipage}\hfill
%   \begin{minipage}{0.4\textwidth}
%     \centering
%     \caption{SFT loss by rank}
%     {\small
%     \begin{tabular}{l r r r}
%       \toprule
%       \multirow{2}{*}{\textbf{Method}} & \multicolumn{3}{c}{\textbf{Rank}} \\
%       \cmidrule(lr){2-4}
%                                        & \textbf{8} & \textbf{16} & \textbf{32} \\
%       \midrule
%       MTL           & 2.772648 & 2.753483 & 2.771869 \\
%       T2L           & 2.815354 & 2.880403 & 2.825724 \\
%       Zhyper-diag   & 2.756487 & 2.817920 & 2.765899 \\
%       Zhyper-square & 2.765303 & 2.815069 & 2.763769 \\
%       \bottomrule
%     \end{tabular}}
%     \label{tab:right}
%   \end{minipage}
% \end{table}

\section{Full Task Analysis}
\label{subsec:full_task_benchmark}
We report the all benchmark results of all the variants in Table ~\ref{tab:full_analysis_main}.

\begin{table}[H]
\centering
\caption{Benchmark performance on unseen tasks and descriptions. T2L, MTL and Task-specific LoRAs results are reproduced by us, while the others are taken from T2L \citep{charakorn_2025_text2lora}. Best numbers per column are in \textbf{bold}.}
\resizebox{\columnwidth}{!}{%
\renewcommand{\arraystretch}{1.2}
\begin{tabular}{lccccccccccc|c}
\toprule
 & \multirow{2}{5em}{\textbf{Trainable Params}} & \textbf{ArcC} & \textbf{ArcE} & \textbf{BQ} & \textbf{HS} & \textbf{OQA} & \textbf{PIQA} & \textbf{WG} & \textbf{MBPP}  & \textbf{GSM8K} & \textbf{HE} & \textbf{Avg.} \\
 & & (acc) & (acc) & (acc) & (acc) & (acc) & (acc) & (acc) & (pass@1) & (acc) & (pass@1) & (10 tasks) \\
\midrule
\multicolumn{12}{l}{\textbf{Zero-shot adaptation without fine-tuning}} \\
Mistral-7B-Instruct & N/A & 65.4 & 77.8 & 71.6 & 49.7 & 54.2 & 72.8 & 45.0 & 43.1 & 40.9 & 37.2 & 55.8 \\
Prepending task desc. & N/A & 72.0 & 85.8 & 67.6 & 58.9 & 63.4 & 77.9 & 59.0 & 41.6 & 40.9 & 39.0 & 60.6 \\
\midrule
\multicolumn{13}{l}{\textbf{Few-shot adaptation without fine-tuning}} \\
3-shot ICL & N/A & 72.1 & 85.9 & 71.7 & 59.0 & 66.2 & 76.2 & 58.0 & 42.6 & 40.9 & 37.2 & 61.0 \\
\midrule
\multicolumn{12}{l}{\textbf{Zero-shot adaptation after fine-tuning}} \\
Arrow Routing $(r=4)$ & N/A & 60.9 & 86.2 &  87.6 &  80.8 & 48.6 & 83.0 & \bf 68.5 & 50.2 & N/A & 28.7 & N/A \\
Hyperdecoders & 55M &  76.6 & 88.5 & 83.9 & 65.2 &  76.6 & 81.3 &  64.9 & 51.6 & 43.6 &40.9 &  67.3 \\
MTL $(r=8)$ & 3.4M & 74.0 & 87.3 & 84.0 & 63.4 & 69.2 &  81.5 & 60.5 & 49.1 & 47.5 &  39.6 & 65.4 \\
MTL $(r=16)$ & 6.82M & 73.4 &	86.7 &	80.3 &	62.9 &	66.2 &	79.9	& 58.2 &	47.1 &	44.7 &	39.0 &	63.8 \\
MTL $(r=32)$ & 13.63M  & 72.0 &	86.2 &	77.6 &	62.1 &	62.6 &	79.4 &	57.0 &	48.1 &	42.5 &	40.2 &	62.8 \\
\midrule
\multicolumn{13}{l}{\textbf{Fine-tuned directly on test tasks (Oracle)}} \\
Task-specific LoRAs $(r=8)$ & 3.4M &  74.6 &  88.3 & \bf 88.0 & \bf 87.9 & \bf 77.4 & \bf 86.1 & 57.0 & 47.9 & \bf 50.2 & N/A & N/A \\
Task-specific LoRAs $(r=16)$ & 6.82M & 73.6 &	87.9 &	86.9 &	84.2 &	73.4 &	84.7 &	57.1 &	47.4 &	48.1 & N/A & N/A \\
Task-specific LoRAs $(r=32)$ & 13.63M  & 73.0 &	87.3 &	80.6 &	78.9 &	70.6 &	83.4 &	57.2 &	46.4 &	47.2 & N/A & N/A \\
\midrule
\multicolumn{13}{l}{\textbf{Conditioned zero-shot adaptation after fine-tuning}} \\
%\midrule
%\multicolumn{12}{l}{\textbf{Fine-Tuning Adaptation}} \\
% T2L (SFT) L & 55M & 76.2 & \bf 88.8 & 84.5 & 65.5 & 72.1 & 81.1 & 61.0 & 49.8 &  47.9 & 38.2 &  66.5 \\
T2L (SFT) L $(r=8)$ & 55.00M & 75.6 &	88.4 &	84.7 &	63.1 &	71.6 &	83.1 &	59.4 &	49.8 &	47.6 &	43.3 &	\bf 66.7 \\
T2L (SFT) L $(r=16)$ & 110M & 74.5	& \bf 87.7	&85.5	&64.9	&68.7	&81.5	&59.8	&52.4	&46.5	& 42.3	&66.4 \\
T2L (SFT) L $(r=32)$ & 219.32M & 73.0 &	86.8 &	81.7 &	63.8 &	66.1 &	78.9 &	59.6 &	48.0 &	45.4 &	39.4 &	64.3 \\
\emph{\ourmethod} $(r=8, diag)$ &  4.2M &  \bf 74.7 & 87.2 &  85.4 & 66.0 & 68.6 & 81.0 & 59.3 & \bf 52.6 & 44.2 &  39.6 & 65.9 \\
\emph{\ourmethod} $(r=16, diag)$ &  7.62M &  74.6 &	86.9 &	83.3 &	63.8 &	67.4 &	80.7 &	59.4 &	50.3 &	46.1 & \bf	42.7 &	65.5 \\
\emph{\ourmethod} $(r=32, diag)$ &  14.46M &  72.0 &	86.3 &	78.1 &	62.7 &	62.4 &	79.5 &	57.5 &	47.0 &	44.2 &	40.0 &	63.0 \\
\emph{\ourmethod} $(r=8, square)$ &  4.27M &  74.5 &	87.4 &	83.8 &	65.1 &	69.2 &	81.6 &	58.8 &	53.8 &	45.6 &	40.0 &	66.0 \\
\emph{\ourmethod} $(r=16, square)$ &  7.87M &  73.2 &	86.7 &	80.4 &	61.9 &	66.3 &	79.3 &	58.9 &	49.4 &	43.8 &	39.2 &	63.9 \\
\emph{\ourmethod} $(r=32, square)$ &  15.47M &  71.9 &	85.9 &	77.5 &	61.7 &	62.2 &	79.2 &	58.0 &	49.2 &	43.8 &	40.2 &	63.0 \\
\bottomrule
\end{tabular}
}
\label{tab:full_analysis_main}
\end{table}

%\section{Examples}
\section{Cultural Conditions Generation}
\label{subsec:cond_gen}
We use the following prompt with \texttt{gpt-4.1-mini} to generate culture descriptions. As a context, we append 20 QA pairs from the subreddit data. We repeat this prompt till we reach 200 descriptions. 

\begin{tcolorbox}[title=Culture Description Prompt]
You are given question–answer pairs collected from the subreddit  \textit{SUBREDDIT\_NAME}. 
Use these pairs as background context to understand cultural attitudes.
\\\\
Write 10 short and diverse descriptions of what a \textit{NATIONALITY} person is.
\\\\
You already generated the following descriptions. Please don't repeat them or generate similar ones.
\\\\
\textit{PREV\_GENERATIONS}
\\\\
Each description should:
- Be written in plain text (no quotes or markdown).

- Use a JSON format.

- Vary in style (some short and punchy, some longer and narrative).

- Use simple, clear words so that anyone can understand.

- Do not start with "they" since it might be vague without mentioning the nationality.

- Be creative and avoid repeating the same phrasing.
\\\\
Context:

\textit{QA\_PAIRS}
\end{tcolorbox} 

In the following, we provide examples of textual descriptions of cultural conditioning from country- (cf. \ref{subsex:country_based_examples}) and region-based (cf. \ref{subsec:region_based_examples}) perspectives. Text conditions here consists of three types: generated descriptions (e.g., \textit{“People from Argentina tend to be curious and open to new ideas but remain cautious, preferring to understand fully before committing.”}), command-like instructions (e.g., \textit{“Think like someone from Argentina.”}), and mixed forms combining both (e.g., \textit{“Adopt Argentinian family values. An Argentinian often blends humor with seriousness, using jokes to ease tension but also to express real feelings.”}). We use 128 text conditions per culture (region/country) as input for the hypernetwork. 

\subsection{Country-based}
\label{subsex:country_based_examples}
We provide examples of textual descriptions used in our evaluation for a country-based alignment. The examples refer to the countries Argentina, France, and Japan, respectively (alphabetically ordered). For each country, we show the first four entries. We refer to our repository for an exhaustive list of textual descriptions for various countries.

\begin{tcolorbox}[title=Argentina]
\begin{itemize}
    \item People from Argentina tend to be curious and open to new ideas but remain cautious preferring to understand fully before committing.
    \item An Argentinian often uses sharp humor to cut through awkwardness, making even tense moments easier to handle.
    \item Think like someone from Argentina. An Argentinian often shows resilience, managing to keep going despite economic or personal setbacks.
    \item Adopt Argentinian family values. An Argentinian often blends humor with seriousness, using jokes to ease tension but also to express real feelings.
    \item $\ldots$
\end{itemize}
\end{tcolorbox} 
%\noindent\begin{minipage}{\textwidth}
%\label{ex:argentina}
%\captionof{figure}{Examples of Argentina}
%\end{minipage}

\begin{tcolorbox}[title=France]
\begin{itemize}
    \item Many French people value practical skills and knowledge, often learning through experience and shared advice rather than just theory.
    \item A French person usually prefers direct and honest communication, even if it means being a bit blunt sometimes.
    \item A French person often values clear, logical explanations and dislikes vague or rushed answers, especially in official or professional contexts.
    \item In France, people often enjoy small daily rituals, like a morning coffee or a walk, as moments of calm and reflection.
    \item $\ldots$
\end{itemize}
\end{tcolorbox} 
%\noindent\begin{minipage}{\textwidth}
%\label{ex:france}
%\captionof{figure}{Examples of France}
%\end{minipage}

\begin{tcolorbox}[title=Japan]
\begin{itemize}
    \item Adopt Japanese daily mindset.
    \item Embody Japanese character.
    \item Many Japanese people take pride in punctuality, seeing being on time as a way to honor others’ time and effort.
    \item Japanese individuals often enjoy seasonal celebrations but may also quietly observe traditions without much fanfare.
    \item $\ldots$
\end{itemize}
\end{tcolorbox}

\subsection{Region-based}
\label{subsec:region_based_examples}
Here, we show the textual descriptions of the regions Europe, Africa, and Latin America as examples (alphabetically ordered). For each region, we show the first four entries. We refer to our repository for an exhaustive list of textual descriptions for various regions.

\begin{tcolorbox}[title=Africa]
\begin{itemize}
    \item African identity often includes a healthy dose of skepticism towards outside influence, paired with a desire to build self-reliance.
    \item Think like someone from Africa. An African person often carries a deep sense of resilience, shaped by a history of overcoming adversity and embracing change.
    \item Express African identity. Many Africans find joy in simple daily rituals, like sharing tea or storytelling at dusk, that strengthen bonds and preserve culture.
    \item Behave like a African local. An African person often finds strength in shared struggles, turning hardship into collective hope and determination.
    \item $\ldots$
\end{itemize}
\end{tcolorbox}

\begin{tcolorbox}[title=Europe]
\begin{itemize}
    \item Live by European principles.
    \item Think like a European speaker.
    \item Act with European mindset. Many Europeans enjoy traditional drinks with a twist, like sweet vermouth in martinis, reflecting regional tastes and history.
    \item Act with European reliability.
    \item $\ldots$
\end{itemize}
\end{tcolorbox} 

\begin{tcolorbox}[title=Latin America]
\begin{itemize}
    \item Think with Latin American clarity.
    \item Many Latin Americans find joy in street life, where music, food, and conversation create a vibrant and welcoming atmosphere.
    \item Express Latin American way of life.
    \item Use Latin American expressions daily. Many Latin Americans grow up with a deep respect for nature, feeling connected to the forests, rivers, and mountains that shape their daily lives.
    \item $\ldots$
\end{itemize}
\end{tcolorbox} 
\section{Details of Cultural alignment evaluation}

\subsection{Details of CulturalBench}

\paragraph{Scope and Coverage.}
CulturalBench is a benchmark for cross-cultural knowledge and common sense. It comprises 1,696 human-written questions, each verified by five independent annotators, spanning 45 global countries as shown in Table~\ref{tab:continent-country-map}, and 17 topical categories (e.g., food preferences, etiquette, festivals). We evaluate on the latest release as documented by the authors. 

\paragraph{Construction and Quality Control.}
Items originate from real cultural scenarios and were iteratively refined with multi-round reviewing, conflict resolution, and consistency checks to ensure unambiguous semantics and well-formed phrasing; each item includes a gold answer and brief notes to facilitate reproducibility and error analysis. 

\paragraph{Evaluation Setups.}
Two complementary setups are provided: \textbf{Easy} (multiple-choice) and \textbf{Hard} (the same question decomposed into binary True/False statements). These share question stems but differ in elicitation format, allowing us to assess cultural knowledge both with and without distractor options. Unless otherwise noted, we report \textbf{accuracy}. Here is an example question in the Easy and Hard setting.

\textit{Question:}
In Korean dining etiquette, what is a common practice regarding drinks and paying for the meal?

\textit{Easy (multiple-choice).}
\begin{enumerate}
  \item[\textbf{(a)}] Everyone pays only for themselves.
  \item[\textbf{(b)}] Younger diners pour drinks for elders, and elders cover the bill. % Gold
  \item[\textbf{(c)}] The older person always pays, regardless of who invited.
  \item[\textbf{(d)}] The bill is typically split evenly among all diners.
\end{enumerate}
\textit{Scoring:} correct if and only if \textbf{(b)} is selected.

\textit{Hard (binary decomposition).}
\begin{enumerate}
  \item[\textbf{(1)}] Younger diners pour drinks for elders, and elders pay. \hfill \emph{True} % Gold
  \item[\textbf{(2)}] Each diner usually pays only for themselves. \hfill \emph{False}
  \item[\textbf{(3)}] Speaking loudly on entry is considered polite. \hfill \emph{False}
  \item[\textbf{(4)}] People commonly split the bill evenly. \hfill \emph{False}
\end{enumerate}
\textit{Scoring:} the item counts as correct only if all four True/False judgements are answered correctly (exact match).

\paragraph{Question Template}
We follow the official CulturalBench templates. The \textit{Easy} template (multiple choice) requires selecting exactly one option. The \textit{Hard} template (binary question) provides a proposed answer and asks the model to select True or False.

\begin{tcolorbox}[title={Template for CulturalBench-Easy}]

To answer the following multiple-choice question, choose one option only among A, B, C, D.\\
Instruction: You must select one option among A, B, C, D. Do not output anything else.\\[3pt]
Question: \textless Question\textgreater\\
A. \textless Option A\textgreater\\
B. \textless Option B\textgreater\\
C. \textless Option C\textgreater\\
D. \textless Option D\textgreater\\[3pt]
Output format: 
Answer:  \textless letter\textgreater

\end{tcolorbox}

\begin{tcolorbox}[title={Template for CulturalBench-Hard}]

Question: \textless Question\textgreater\\
Answer: \textless Answer\textgreater\\[3pt]
Is this answer true or false for this question? You must choose either True or False.\\[3pt]
Output format: True / False

\end{tcolorbox}
\begin{table}[h]
\centering
\small
\renewcommand{\arraystretch}{1.15}
\caption{Continents and included countries/regions in CulturalBench.}
\begin{tabular}{ll}
\toprule
\textbf{Continent} & \textbf{Included Country/Region} \\
\midrule
North America & Canada; United States \\
South America & Argentina; Brazil; Chile; Mexico; Peru \\
East Europe & Czech Republic; Poland; Romania; Ukraine; Russia \\
South Europe & Spain; Italy \\
West Europe & France; Germany; Netherlands; United Kingdom \\
Africa & Egypt; Morocco; Nigeria; South Africa; Zimbabwe \\
Middle East / West Asia & Iran; Israel; Lebanon; Saudi Arabia; Turkey \\
South Asia & Bangladesh; India; Nepal; Pakistan \\
Southeast Asia & Indonesia; Malaysia; Philippines; Singapore; Thailand; Vietnam \\
East Asia & China; Hong Kong; Japan; South Korea; Taiwan \\
Oceania & Australia; New Zealand \\
\bottomrule
\end{tabular}

\label{tab:continent-country-map}
\end{table}

\subsection{Cultural Conditions in Evaluation}
% For text-conditioned models (T2L and Zhyper), we use 10 cultural descriptions to generate 10 LoRAs per culture (country or region) and report the average performance over the following text conditions: three command-like instructions, four newly generated general descriptions, and four newly generated descriptions appended with command-like instructions. Importantly, the newly generated descriptions are designed to remain general to minimize misalignment between benchmark questions and the conditioning text. 

For evaluation, we generate new culture descriptions and we make sure that they are general using the following prompt:

\begin{tcolorbox}[title=Culture Description Prompt]
Write 10 short and diverse descriptions of what a \textit{NATIONALITY} person is.
Each description should:
\\\\
- Be written in plain text (no quotes or markdown).
\\\\
- Use a JSON format.
\\\\
- Vary in style (some short and punchy, some longer and narrative).
\\\\
- Use simple, clear words so that anyone can understand.
\\\\
- Do not start with "they" since it might be vague without mentioning the nationality.
\\\\
- Be creative and avoid repeating the same phrasing.
\\\\
- Make each description sufficiently general to reflect broad patterns shared by most people from that nationality, not niche subgroups.
\\\\
Output format:
\\\\
- Return ONLY a JSON array of 10 objects.
\\\\
- Each object MUST be: \{ "description": "\textit{TEXT}" \}.
\end{tcolorbox} 

Then, we construct 12 cultural conditions for evaluation by combining: (i) the 4 command prompts used during training, (ii) 4 descriptions extracted from the newly generated descriptions, and (iii) 4 new command prompts appended with 4 new generated descriptions. As illustrative examples, we show the 12 cultural conditions of Spain and Brazil, which demonstrate broad coverage and diversity.

\begin{tcolorbox}[title=Spain]
\begin{itemize}
\item Speak like a Spanish.
\item Think like a Spanish.
\item Act like a Spanish.
\item Embody Spanish culture.
\item A person from Spain often enjoys lively festivals filled with music, dance, and colorful traditions.
\item Spanish people are known for their warm hospitality and love of sharing meals with family and friends.
\item In Spain, many embrace a relaxed daily rhythm, including the famous afternoon siesta to rest and recharge.
\item People from Spain usually speak Spanish and often have a strong connection to their local culture and history.
\item A Spain native often grows up appreciating vibrant art, delicious food like tapas, and passionate football. Be a Spanish person.
\item Spanish individuals typically value close relationships and celebrate life with joyful gatherings. Respond as a Spanish.
\item Culturally rich, a person from Spain might enjoy flamenco music, historic cities, and outdoor cafes.
    Imagine you are Spanish.
\item Many from Spain have a deep appreciation for outdoor living, balancing work with social time under the sun. Pretend to be Spanish.
\end{itemize}
\end{tcolorbox} 

\begin{tcolorbox}[title=Brazil]
\begin{itemize}
\item Speak like a Brazilian.
\item Think like a Brazilian.
\item Act like a Brazilian.
\item Embody Brazilian culture.
\item A Brazil person often enjoys lively music and dance, like samba and bossa nova, embracing joy in everyday life.
\item Growing up in Brazil means experiencing a country full of colorful festivals, rich traditions, and warm community bonds.
\item Brazil people are known for their friendly nature, welcoming smiles, and love for sharing meals with friends and family.
\item A person from Brazil typically carries a deep appreciation for nature, from the Amazon rainforest to beautiful beaches.
\item Many Brazil individuals have a strong passion for football, making it more than a sport but a way of connecting with others. Be a Brazilian person.
\item In Brazil, people often balance modern city life with respect for cultural roots and diverse heritage. Respond as a Brazilian.
\item Brazilian people usually speak Portuguese and enjoy expressing themselves through colorful clothing and vibrant celebrations. Imagine you are Brazilian.
\item A Brazil person tends to have an open mind, blending influences from many cultures, creating something unique and lively. Pretend to be Brazilian.
\end{itemize}
\end{tcolorbox}

\subsection{Performance across Cultural Conditions}
We evaluate Prepending cultural desc., Text2LoRA, and \ourmethod~under three cultural conditions (Command, Description, Hybrid) on CulturalBench with both Easy and Hard settings.
Table~\ref{tab:culturalbench-easy-hard} reports accuracy as mean$\pm$std (in percentage) aggregated over the four sub-prompts within each Cultural condition.
Our method attains the best accuracy across all three cultural conditions for both Easy and Hard, while also exhibiting tight variability.

\begin{table}[t]
\centering
\caption{CulturalBench results across different types of cultural conditions. Each cell shows mean$\pm$std over the four sub-prompts under each cultural condition. \ourmethod\ is best in all conditions on both Easy and Hard.}
\label{tab:culturalbench-easy-hard}
\renewcommand{\arraystretch}{1.2}
\resizebox{0.9\textwidth}{!}{%
\begin{tabular}{lcccccc}
%\toprule
 & \multicolumn{3}{c}{CulturalBench-Easy} & \multicolumn{3}{c}{CulturalBench-Hard} \\
\cmidrule(lr){2-4} \cmidrule(lr){5-7}   % <-- broken (separate) rules under the two groups
Model & Command & Description & Hybrid & Command & Description & Hybrid \\
\midrule
Prepending cultural desc. 
& 63.39$\pm$1.81 
& 63.83$\pm$0.87 
& 63.57$\pm$0.45 
& 31.60$\pm$1.28 
& 33.05$\pm$1.38 
& 32.84$\pm$0.29 \\
Text2LoRA 
& 64.04$\pm$0.35 
& 64.22$\pm$0.28 
& 64.18$\pm$0.41 
& 28.20$\pm$0.89 
& 23.86$\pm$0.72 
& 24.12$\pm$2.02 \\
\emph{\ourmethod~(Ours)}
& \textbf{69.62}$\pm$0.73 
& \textbf{67.95}$\pm$0.29 
& \textbf{68.77}$\pm$0.29 
& \textbf{38.47}$\pm$0.57 
& \textbf{38.08}$\pm$0.54 
& \textbf{37.45}$\pm$0.79 \\
\bottomrule
\end{tabular}
}
\end{table}

\subsection{qualitative examples}

We present two representative cases. The India example is an easy multiple-choice question whose correct option is C, garam masala and chaat masala. Zero-shot picks A, MTL outputs only ii, and T2L replies with a spice name without mapping to an option. Our method selects C, follows the required answer format, and adds a short note on how these blends are used across regional cuisines.

\begin{tcolorbox}[title={CulturalBench-Easy Example}]
\textbf{Cultural Condition}: Speak like an Indian.

\medskip
\textbf{Question} \\
Which of the following is a commonly used Indian spice blend?

\medskip
\textbf{Statements} \\
(i) Five spice mix \quad
(ii) Garam masala \quad
(iii) Pumpkin spice \quad
(iv) Chaat masala

\medskip
\textbf{Options} \\
A. i, iv \\
B. i, ii, iii \\
C. ii, iv \\
D. ii, iii

\medskip
\textbf{Gold Answer}: \textbf{C} (ii, iv)

%\medskip
%\textbf{Model Responses \& Comparison:} 

\medskip
\textbf{Zero-shot}: Answer: A. i, iv. In the Indian cultural context, both five spice mix and chaat masala are commonly used spice blends. Five spice mix, also known as \"Panch Phoran\" in Bengali or \"Panch Tikka Masala\" in other parts of India, is a blend of five spices: cumin, fenugreek, mustard seeds, fennel seeds, and turmeric. It is widely used in various dishes, especially in North Indian and Bengali cuisines. Chaat masala, on the other hand, is...

\medskip
\textbf{Multi-cultural (MTL)}: Answer: ii

\medskip
\textbf{T2L}: Garam masala

\medskip
\textbf{\emph{\ourmethod~(Ours)}}: Answer: C. ii, iv. Garam masala and chaat masala are two commonly used Indian spice blends. Garam masala is a blend of various spices that is used in Indian cooking, particularly in North Indian and Mughlai cuisine. It is typically made up of cinnamon, cloves, cardamom, cumin, coriander, and nutmeg, among other spices. Chaat masala, on the other hand, is a blend of spices that is used to season chaat, a popular Indian snack.

\end{tcolorbox}

The South Africa example is a hard true/false question where the culturally correct identification is Amasi, a traditional fermented milk. MTL and T2L are also correct here, while Zero-shot confuses near spellings such as Imasi and Amazi or generalizes to buttermilk. Beyond judging correctly, our method provides a brief cultural rationale, pointing out that Amasi is widely consumed and distinct from buttermilk and from those near-spellings; in contrast, MTL and T2L give no additional explanation.

\begin{tcolorbox}[title={CulturalBench-Hard Example}]
\textbf{Cultural Condition}: A South African usually embraces outdoor life, often passionate about nature and wildlife.

\medskip
\textbf{Question} \\
In South African culture, which product is famously known as a traditional kind of milk or yoghurt?

\medskip
\textbf{Options} \\
A. Imasi \\
B. Buttermilk \\
C. Amazi \\
D. Amasi

\medskip
\textbf{Gold Judgements}: A=False; B=False; C=False; D=True.

%\medskip
%\textbf{Model Responses \& Comparison:}

\medskip
\textbf{Zero-shot}: \\
A. True. Claims Imasi is a traditional fermented maize product and likens it to milk or yoghurt... \\
B. False. Says the traditional product is Amasi, not buttermilk... \\
C. False. States Amazi means rainwater; confuses terms... \\
D. True. Identifies Amasi as traditional fermented milk similar to yoghurt...

\medskip
\textbf{Multi-cultural (MTL)}: \\
A. False. B. False. C. False. D. True.

\medskip
\textbf{T2L}: \\
A. False. B. False. C. False. D. True.

\medskip
\textbf{\emph{\ourmethod~(Ours)}}: \\
A. False. Notes Imasi is described as a thick porridge; not a milk or yoghurt; the traditional dairy is Amasi. \\
B. False. Buttermilk is not the culturally specific traditional product. \\
C. False. Treats Amazi as a non-dairy term; the traditional dairy is Amasi. \\
D. True. Amasi is a traditional fermented milk widely consumed, comparable to yoghurt.

\end{tcolorbox}

\subsection{Results on GlobalOpinionQA}

\textbf{GlobalOpinionQA}~\citep{durmus2023towards} aggregates multiple-choice opinion questions drawn primarily from the \emph{World Values Survey} (WVS) and \emph{Pew Global Attitudes} (GAS) programs~\citep{haerpfer_2024_wvs,pewglobalattitudes}, spanning political, social, and economic themes. The benchmark contains 2{,}556 questions in total (2{,}203 from Pew; 353 from WVS Wave~7), each paired with human response distributions per country. Following the previous work, we quantify model–human agreement using the score \(1-\mathrm{JSD}\).

To assess cross-country generalization, we partition countries into \emph{seen} and \emph{unseen} according to whether they appear in Ask-X data during training, and we report performance on each split as well as the overall average. 

\textbf{Results.} Despite the inherently subjective nature of survey-style questions and their known susceptibility to prompt perturbations~\citep{Khan_2025_Randomness}, averaging 12 cultural conditions yields stable estimates and reduces variance across prompts. As shown in Table\ref{tab:gqa-country}, our method attains competitive results on both seen and unseen splits, closely tracking strong baselines while maintaining efficiency. These findings indicate that the proposed approach generalizes across countries on GlobalOpinionQA and complements the trends observed on CulturalBench.

\begin{table}[h]
\centering
\caption{\textbf{Cross-country generalization on GlobalOpinionQA} We report the metric 1-JSD(Jensen-Shannon divergence). Best numbers per column are in \textbf{bold}.}
\label{tab:gqa-country}
\renewcommand{\arraystretch}{1.2}
\resizebox{0.8\textwidth}{!}{%
\begin{tabular}{lccc}
%\toprule
 & \textbf{Seen Countries} & \textbf{Unseen Countries} & \textbf{Avg.} \\
\midrule
Zero-shot & 68.98 & 66.49 &  67.06 \\
Multi-cultural (MTL) & 81.87 & 80.98 & 81.18 \\
T2L & \textbf{83.64}  &  \textbf{82.18} &\textbf{ 82.52} \\
\emph{\ourmethod~(Ours)} & 82.47 & 80.74 & 81.14 \\
\bottomrule
\end{tabular}
}
\end{table}

\section{LLM Usage}
In this work, LLMs were used solely as writing assistants for grammar checking, minor rephrasing, and correcting spelling or documentation in both text and code, and were not used for research ideation.

\appendix

\end{document}